\documentclass[12pt]{article}

\usepackage{amsmath,amsthm, amsfonts, amssymb, amsxtra, amsopn}
\usepackage{pgfplots}
\pgfplotsset{compat=1.13}
\usepackage{pgfplotstable}
\usepackage{graphicx,grffile}
\usepackage{multirow}
\usepackage{booktabs}
\usepackage{listings}
\usepackage{cmap}
\usepackage{colortbl}
\usepackage{adjustbox}
\usepackage{epsfig}

\usepackage[tableposition=top,font=small,skip=5pt]{caption}
\usepackage{makecell} 

\def\calN{{\cal N}}

\PassOptionsToPackage{hyphens}{url}

\usepackage{hyperref}
\hypersetup{colorlinks=true,linkcolor=black,citecolor=black,urlcolor=blue,filecolor=black}
\hypersetup{pdfpagemode=UseNone,pdfstartview=}

\usepackage{enumitem}
\setlist[itemize]{noitemsep, topsep=0pt}

\advance\oddsidemargin by -0.35in
\advance\textwidth by 0.7in

\advance\topmargin by -0.4in
\advance\textheight by 0.8in

\def\eref#1{(\ref{#1})}

\long\def\symbolfootnotetext[#1]#2{\begingroup%
\def\thefootnote{\fnsymbol{footnote}}\footnotetext[#1]{#2}\endgroup}

\def\zz{\phantom{0}}

\title{A Comparison of Graph Neural Networks for Malware Classification}

\author{Vrinda Malhotra\footnotemark[1]\ \ \ 
Katerina Potika\footnotemark[1]\ \ \
Mark Stamp\footnotemark[1]\,\,\footnotemark[2]}

\begin{document}

\symbolfootnotetext[1]{Department of Computer Science, San Jose State University}
\symbolfootnotetext[2]{mark.stamp$@$sjsu.edu}

\maketitle

\abstract
Managing the threat posed by malware requires
accurate detection and classification techniques. Traditional detection strategies, such as signature 
scanning, rely on manual analysis of malware to extract relevant features, which is labor intensive 
and requires expert knowledge. Function call graphs consist of a set of program functions and their 
inter-procedural calls, providing a rich source of information that can be leveraged to classify malware 
without the labor intensive feature extraction step of traditional techniques. In this research, 
we treat malware classification as a graph classification problem. Based on Local Degree Profile 
features, we train a wide range of Graph Neural Network (GNN) architectures to generate embeddings 
which we then classify. 
We find that our best GNN models 
outperform previous comparable research involving the well-known 
MalNet-Tiny Android malware dataset. In addition, our GNN models do not 
suffer from the overfitting issues that commonly afflict non-GNN techniques,
although GNN models require longer training times.

\section{Introduction}\label{chap:1}

Android malware is a common problem on mobile devices and poses a serious challenge due to its 
volume and diversity. According to AV-TEST~\cite{bib1}, in~2021 some~9.09 million 
new mobile malware samples were intercepted, which is an average of nearly~25,000 new samples per day.
Of these mobile malware samples, about~3.5 million were Android-based. 

Since Android is an open-source platform, it provides flexibility for mobile developers to create custom applications (apps). 
However, this same flexibility can be exploited by bad actors to create malicious apps that users install. 
Therefore, solutions that detect and classify malware are 
crucial for the safety of Android devices. 

Previous studies such as~\cite{bib32} have illustrated that new malware apps tend to be 
variants of pre-existing malware families,
and therefore showcase similar behavioral traits. For example, 
the GinMaster malware family sends confidential information to a remote server, 
while the DroidKungFu family allows a hacker to control an infected device from a 
remote location and access confidential information. 
We can classify malware samples into broad families based on their
differing characteristics~\cite{bib30}.

Signature-based malware detection methods have been 
popular for creating endpoint protection systems because they are 
comparatively fast and effective on ``traditional'' malware. 
However, generating classic signatures requires expert knowledge 
to reverse engineer malware instances and produce the necessary features, 
and hence reverse engineering cannot scale with malware production. 
Signature-based techniques also fail to identify zero-day malware, 
whereas machine learning-based methods have the potential to identify 
and classify previously unseen malware~\cite{bib31}.

Static malware analysis relies on features that can be extracted without executing or emulating code,
in contrast to dynamic analysis, where execution or emulation is required~\cite{anusha}. In general,
static analysis is more vulnerable to code obfuscation techniques employed by modern
polymorphic and metamorphic malware~\cite{bib63}, while dynamic malware analysis 
is more resistant to such obfuscations~\cite{bib61}. 
However, collecting dynamic features is more resource and time intensive~\cite{bib62}. 

Studies such as~\cite{bib32} perform broad static analysis to generate features (e.g., user permissions, 
suspicious API calls, network addresses, etc.) and then use machine learning algorithms to classify malware. 
However, these machine-learning algorithms generally do not adequately model interactions between function calls. 
Such interactions can most naturally be taken into account using a graph-like data structure. Graph-based methods do not 
assume that the features of a particular instance are independent of each other, and the models themselves 
can learn relationships between features. This provides us with an additional layer of information that is inherent 
in the input data and can be best utilized by graph models. Moreover, labor-intensive manual analysis of code is not required 
to generate feature sets for our graph-based models, giving them a significant advantage over traditional techniques, 
such as signature analysis. The goal of this research is to explore malware classification techniques using graph-based 
learning that relies solely on graphs generated from code.

The remainder of this paper is organized as follows.
In Section~\ref{chap:2}, we discuss relevant related work. In Section~\ref{chap:3}, 
we introduce several learning techniques that are based on different principles, ranging from standard machine
learning models to current state-of-the-art Graph Neural Networks (GNN). 
Section~\ref{chap:4} includes implementation details related to the 
experiments that we perform. In Section~\ref{chap:5}, we analyze and discuss our experimental results. 
Finally, in Section~\ref{chap:6}, we provide a summary of our main results and discuss potential directions
for future work.

\section{Related Works}\label{chap:2}

In this section, we highlight some of the learning-based methods 
that have been previously used for malware classification and detection. 
We divide this selective survey between traditional learning-based classifiers and graph-based classification.
Under the heading of ``traditional'' learning techniques, we include both classic 
statistical-based machine learning and neural networking-based deep learning.

\subsection{Traditional Learning-Based Classification}

In the paper~\cite{bib34} from 2013, a Bayesian classifier was trained to detect malware using~58 
code-based attributes. In the 2014 paper that introduced the popular Drebin dataset~\cite{bib32}, 
features (e.g., permissions and API calls) were extracted from Android \texttt{apk} files
and the samples were then classified using Support Vector Machines (SVM).  
A later method, which its authors referred to as Significant Permission Identification 
for \hbox{Android} Malware Detection (SIGPID)~\cite{bib36}, mined the permissions data of 
each malware app using three levels of pruning to identify the~22 most significant permissions.
These features were then used to train an SVM classifier. 

With the rise in popularity of deep learning models, 
papers such as~\cite{bib39} appeared, where permission sequences were extracted as word embeddings and 
used as features in a Long Short-Term Memory Network (LSTM) model. Similarly, in~\cite{bib31}, 
API features were selected based on their frequency and then mapped to an image-like structure,
with Convolutional Neural Networks (CNN) used as the classifier. The results indicate
API calls and permissions are strong features for Android malware detection, 
but that the optimal choice of these features depends on the dataset used. 

Opcode-sequence $n$-gram features have also been successfully used to detect malware~\cite{bib35}. 
In addition, the paper~\cite{bib38} used raw opcode sequences encoded as one-hot vectors, 
which served to classified samples via a CNN model. Techniques based on opcode features
can be defeated, since opcode sequences can be directly altered via elementary obfuscation
techniques. Another popular approach for malware classification is to convert 
files into grayscale images and then take advantage of the strength of CNNs for malware 
detection and classification~\cite{bib60,bib59}.
Additional work on malware classification includes a range of novel approaches, 
such as that in~\cite{bib37}, where patterns based on inter-component communication
were extracted from the code. 

\subsection{Graph Learning-Based Classification}

While traditional learning techniques are based on the type of feature and classifier, 
graph-based learning relies on the type of graph structure and how the node features are embedded. 
For example, the papers~\cite{bib4,bib6,bib9} use API call graphs. Specifically,
in~\cite{bib6}, apps are represented in relation to APIs, and API relationships are 
mapped as a structured Heterogeneous Information Network (HIN). 
Then a meta-path-based approach is used to characterize the semantic relatedness of apps and APIs, 
aggregating similarities using multi-kernel learning. Heterogeneous graphs are those that
use more than one type of relationship to make connections between nodes. 

In~\cite{bib9}, API call sequence graphs 
learn multiple embedding representations that are successfully used for malware detection and classification. 
The authors use a recurrent neural network to decode the deep semantic information and to independently extract features, 
with a version of a Graph Convolutional Network (GCN) used to model high-level graphical semantics. 
In~\cite{bib4}, a skip-gram model is used to extract features for graph nodes based on API sequences. 
These features rely on both app-API relationships and API-API relationships to form a heterogeneous graph.

Another popular type of graph used for malware analysis is an opcode-level 
function call graphs~\cite{bib12, bib11, bib8}. The research in~\cite{bib12} uses 
opcode-sequences as text features, with traditional machine learning 
methods (Random Forest,  SVM, etc.) for classification. In contrast, the paper~\cite{bib8} 
relies on an LSTM-based neural network for classification. The authors of~\cite{bib11} 
utilize a novel graph structure that they call a co-opcode graph, which is constructed by extracting 
metamorphic engine-specific opcode patterns. A method related to Hidden Markov Models (HMM) 
is then used to classify the malware samples. 

Opcode-level graphs can also be treated like 
text features rather than graphical data. For example, in~\cite{bib5} dynamically-generated network flow graphs
are used to create  a new model that its authors call Network Flow Graph Neural Network (NF-GNN).
This NF-GNN model relies on a novel edge feature-based GNN for classification. 

The graph methods mentioned above are all transductive, which implies that they cannot be
expected to generalize to zero-day malware. One goal of our research is to analyze graph-based models 
that are inductive, that is, models that can be used to predict zero-day malware.
We will base our models on Control Flow Graphs (CFG). In a CFG, nodes represent program statements, 
including called subroutines and conditionals, while edges represent the flow of the program. 
CFGs intra-procedural mappings may be well-suited to classify zero-day malware~\cite{bib7}, 
since they only rely on characteristics of the code of an \texttt{apk}. In~\cite{bib7}, a 
Deep Graph Convolutional Neural Network (DGCNN) is used to embed structural 
information inherent in CFGs for effective and efficient malware classification. 

The inter-procedural counterpart to CFGs are Function Call Graphs (FCG),
where nodes represent subroutines and edges represent the caller-called relationships between
subroutines. In~\cite{bib8}, opcode level function call graphs are obtained from the static analysis of 
malware, while~\cite{bib14} uses a Natural Language Processing (NLP)-inspired graph embedding to 
convert the graph structure of an Android app into a vector. On a related note, the paper~\cite{bib21} 
introduced the MalNet dataset, a large Android malware dataset, where FCG data has been 
extracted from \texttt{apk} files. The authors considered state-of-the-art graph representation learning 
approaches, including GraphSAGE~\cite{bib18} and Graph Isomorphism Networks (GIN)~\cite{bib22}. 
Among the methods considered, they found that FEATHER and GIN achieved the highest classification 
performance. Using this same MalNet dataset, the authors of~\cite{bib57} employ 
Jumping Knowledge GNNs (JK-GNN) with node features 
set to page rank, degree nodes, etc. They conclude that JK-GraphSAGE performs best. 

\subsection{Datasets}\label{sect:datasets}

In our research, we want to focus on graphs that do not require 
handcrafted features, graphs that will enable us to deal with zero-day malware, and graphs that are 
straightforward to generate and analyze. Network flow graphs and control flow graphs are difficult to 
generate and can require significant resources to store, while heterogeneous graphs are difficult to 
interpret and encode. In contrast, a Function Call Graph (FCG) 
is easy to interpret since it consists of all possible execution paths called during runtime. 
For example, if an app is making a sequence of function calls to gather 
sensitive information and send it to a server, all of these function calls are in the FCG,
and we should be able to gain insight into this malicious behavior from the FCG.
Henceforth, we focus on FCGs in this research. 

A comparison between popular
Android malware datasets is provided in Table~\ref{tab:1}. 
Not mentioned in Table~\ref{tab:1} is the AndroZoo dataset~\cite{bib23},
which is an ongoing project of that at the time of this writing
consists of more than~21,000,000 \texttt{apk} files.
The MalNet-Tiny dataset is derived from AndroZoo.

\begin{table}[!htb]
\caption{Popular datasets for Android malware classification}\label{tab:1}
\centering\def\zzzz{\phantom{1,26}}
\adjustbox{scale=0.85}{
\begin{tabular}{c|c|c|c}\midrule\midrule
Dataset & Families & Samples & Papers\\ \midrule
Android Malware Dataset (AMD) & \zz42 & \zzzz5000 & \cite{bib4,bib9,bib14} \\
Android Malware Genome (AMG) & \zz72 & \zzzz1260 & \cite{bib14} \\
Drebin Dataset & 179 & \zzzz5560 & \cite{bib4,bib9,bib14} \\
MalNet & 696 & 1,262,024 & \cite{MalNetWeb} \\
MalNet-Tiny & \zz\zz5 & \zzzz5000 & \cite{bib23,bib21}\\
\midrule\midrule
\end{tabular}
}
\end{table}

All of the datasets in Table~\ref{tab:1} consist of malware \texttt{apk} files or hexadecimal representations 
of the binary content, along with manifests that contain metadata, such as function calls, strings, etc. 

For our experiments, we use the MalNet-Tiny dataset.
As noted in Table~\ref{tab:1}, MalNet-Tiny is comprised of five families,
with~5000 samples in total. 
This dataset has a sufficient number of samples 
to train the graph-based learning models that we consider and, furthermore, each sample
is provided in the form of an FCG~\cite{bib21}. 
Also, MalNet-Tiny has been used in previous studies, 
enabling a direct comparison of our research to previous work.

\subsection{Related Work}

Table~\ref{tab:B9} summarizes a selection of previous work where either traditional learning techniques
or graph-based learning has been applied to the Android malware detection or classification
problems. Note that two of the research papers in Table~\ref{tab:B9} use the MalNet-Tiny
dataset, and hence these two are the most directly relevant for comparison to the results
that we provide in Sections~\ref{chap:4} and~\ref{chap:5}, below.

\begin{table}[!htb]
\caption{Summary of selected previous work}\label{tab:B9}
\centering
\adjustbox{scale=0.6}{
\begin{tabular}{c|ccccc}
\midrule\midrule
\multirow{2}{*}{Paper} & Learning & \multirow{2}{*}{Feature(s)} & Classification & \multirow{2}{*}{Dataset(s)} & Accuracy \\
          & technique(s) & & or detection &  & or F1-score \\ \midrule
Arp, et al.~\cite{bib32} & SVM & Multiple & Classification & Drebin & 0.9400 \\
Busch, et al.~\cite{bib5} & NF-GNN & Network flows & Classification & Custom dataset & 0.9614 \\
Freitas and Dong~\cite{bib21} & GIN & FCG & Classification & MalNet-Tiny & 0.9000 \\
Gao, et al.~\cite{bib4} & GCN & API calls & Both & AMG, Drebin, AMD & 0.9504 \\
G\"{u}lmez and Sogukpinar~\cite{bib12} & RF and SVM & Opcodes & Detection & Custom dataset & 0.9700 \\
He, et al.~\cite{bib60} & ResNet & Images & Detection & Andro-dumpsys & 0.9500 \\
Hou, et al.~\cite{bib6} & HIN & API calls & Detection & Comodo Cloud & --- \\
Huang, et al.~\cite{bib31} & CNN & API calls & Detection & Custom dataset & 0.9430 \\
Jerome, et al.~\cite{bib35} & SVM & Opcode $n$-grams & Classification & AMG plus & 0.8931 \\
Kakisim, et al.~\cite{bib11} & Custom & Co-opcode graph & Detection & Custom dataset  & 0.9930 \\
Khan, et al.~\cite{bib59} & ResNet/GoogleNet & Opcodes & Detection & MMC & 0.8836 \\
Li, et al.~\cite{bib36} & Custom & Permissions & Detection & Custom dataset & 0.9362 \\
Lo, et al.~\cite{bib57} & JK & FCG & Classification & MalNet-Tiny & 0.9440 \\
McLaughlin, et al.~\cite{bib38} & CNN & Opcodes & Classification & AMG plus & 0.8600 \\
Niu, et al.~\cite{bib8} & LSTM & Opcodes & Classification & Custom dataset & 0.9700 \\
Pei, et al.~\cite{bib9} & Graphs & API/permissions & Detection & Drebin, AMD & 0.9967 \\
Vinayakumar, et al.~\cite{bib39} & LSTM & API calls & Classification & AMG plus & 0.8970 \\
Xu, et al.~\cite{bib37} & Custom & ICC & Detection & Custom dataset & 0.9740 \\
Xu, et al.~\cite{bib14} & GNN & FCG & Classification & AndroZoo, Drebin, AMD & 0.9933 \\
Yan, et al.~\cite{bib7} & DGCNN & CFG & Classification & MMC & 0.9942 \\
Yerima, et al.~\cite{bib34} & Bayesian & Multiple & Classification & Custom dataset & 0.8450 \\
\midrule\midrule
\end{tabular}
}
\end{table}

\section{Background}\label{chap:3}

A graph is a data structure consisting of two components, namely, 
nodes (or vertices) and edges. A graph~$G$ is defined as~$G = (V, E)$, 
where~$V$ is the set of nodes, and~$E$ is the edges between them. 
These edges can be directed or undirected. 

Conventional machine learning 
and deep learning techniques can be viewed as dealing with 
relatively simple types of graphs. For example, images can be viewed 
grid graphs while text and speech are sequential and can be viewed as line graphs. However, in general, 
graphs can be far more varied and complex, with arbitrary connections between nodes. 

Recently, analyzing graphs with machine learning has become popular. Traditional feedforward
machine learning algorithms, such as Multi-Layer Perceptrons (MLP) and CNNs, treat the features 
of a particular instance as if they are independent of other instances. While Recurrent Neural Networks (RNN) 
can deal with sequential data, Graph Neural Networks (GNN) allow us to model arbitrary interactions that 
are beyond the scope of RNNs.

Graph Neural Networks (GNN) are neural networks that directly operate on graphs~\cite{bib2}. 
GNNs can be viewed as generalizations of CNNs that allow for a richer neighborhood structure. 
Thus GNNs can model more complex input and output relationships. 
%

\subsection{LDP Features}

In our GNN models, we use the Local Degree Profile (LDP)~\cite{bib33} for node features. LDP is a relatively
simple representation scheme that is based on a node and its one-hop neighborhood. Specifically, 
for a graph~$G=(V,E)$, let~$d(v)$ be the degree of~$v$, that is, the number of vertices that are
adjacent to~$v$ in the graph~$G$. The LDP of a given node~$v\in V$ is defined as the vector 
of length five given by
\begin{equation}\label{eq:LDP}
   \mbox{LDP}(v) = \Big( d(v), \min \big(N(v)\big) , \max \big(N(v)\big), \mu \big(N(v)\big), \sigma \big(N(v)\big) \Big)
\end{equation}
where~$N(v) = \{d(u)\,|\,(v,u)\in E\}$, $\mu$ is the mean,  and~$\sigma$ is the standard deviation~\cite{bib33}.
Thus the LDP of~$v$ is a statistical profile of the neighborhood structure of~$v$ within the graph~$G$.
LDP features have been shown to perform well with the graph neural networks that we discuss
in Section~\ref{sect:GNN-basic}, below.
For our purposes, one potential advantage of the LDP in~\eref{eq:LDP} is that
standard deep learning techniques can also be trained using this feature. 
We note in passing that graph kernel methods reduce dimensionality of the graph data, resulting in 
different features from LDP, while random walk and spectral distance-based methods also 
learn representations of the graph data that differ from the LDP.

Next, we outline each of the learning methods that we consider in this research. 
We divide the models into two groups, namely, non-GNN and GNN models.

\subsection{Non-GNN Learning Methods}\label{sect:non-GNN-learning}

The learning methods in this section use graph data for classification, but
are not considered GNNs. These five architectures provide a baseline for
comparing more costly---with respect to training---GNN architectures.

\subsubsection{Multi-Layer Perceptron}

Multi-Layer Perceptron (MLP) is a standard 
feedforward deep learning technique.
MLPs have proven extremely useful in a wide range of applications.
Here, we train an MLP on the LDP features in~\eref{eq:LDP}, 
which serves as a baseline for comparison of
the effectiveness of our graph-based models. 
%

\subsubsection{Weisfeiler-Lehman Kernel}\label{sect:WL}

Graph kernels methods are a form of supervised learning that use kernel functions to reduce 
dimensionality.  In contrast to traditional kernel-based machine learning models, 
such as Support Vector Machines (SVM), graph kernel methods are designed 
specifically to deal with graph-based features.

In our experiments, we use the Python GraKel library to test the 
Weisfeiler-Lehman Subtree Kernel (WL-Kernel) technique~\cite{bib26}. 
Such models can be used, for example, to find isomorphisms between graphs~\cite{bib25}. 

\subsubsection{FEATHER}

FEATHER~\cite{bib40} is a complex representation scheme that uses characteristic functions of node features 
with random walk weights to describe node neighborhoods and to create node embeddings. 
FEATHER introduces a generalization of characteristic functions to node neighborhoods, 
where the probability weights of the characteristic function are defined by a measure known
as ``tie strength.'' The so-called $r$-scale random walk weighted 
characteristic function is then used to generate the embeddings. 
Random Forest is used to classify based on the resulting embeddings.

\subsubsection{Slaq-VNGE and Slaq-LSD}

We experiment with Slaq-VNGE and Slaq-LSD, both of which approximate the spectral distances 
between graphs, but are based on different functions. Whereas Slaq-VNGE uses 
Von Neumann Graph Entropy (VNGE), which measures information divergence and 
distance between graphs~\cite{bib41}, Slaq-LSD approximates NetLSD, 
which measures the spectral distance between graphs based on 
the heat kernel~\cite{bib42}.

\subsection{GNN Architectures}\label{sect:GNN-basic}

In this section, we consider GNN architectures. 
In this section, we consider networks that attempt to improve on the basic 
GNN architectures of the previous section. These models are more complex
and costly to train, as compared to the GNNs discussed above.

\subsubsection{Graph Convolutional Networks}

Convolutional Neural Networks (CNN) have been extremely successful for image classification. 
However, CNNs are limited to simple grid-like graph structures. GNNs can be viewed as generalizations 
of CNNs to more general graph structures. GNNs aim to generate node embeddings that transform the graph nodes 
into a low-dimensional embedding space. The mean of all node embeddings is taken to form the whole 
graph embedding, which encodes the whole graph into low-dimensional space for graph classification~\cite{bib27}.  


Graph Convolutional Networks (GCN) compute node embeddings by aggregating neighborhood node 
features. Let~$A$ be the adjacency matrix of the graph~$G=(V,E)$.
Then~$A=\{a_{ij}\}$ is an~$N \times N$ matrix, where~$N=|V|$ and~$a_{ij}$ is~1 if~$(v_i,v_j)\in E$; 
otherwise~$a_{ij}=0$. Each node has a $k$-dimensional feature vector, with~$X \in R^{N \times k} $ 
representing the feature matrix for all~$N$ nodes. An~$L$-layer GCN consists of~$L$ 
graph convolution layers, with each layer constructing embeddings for every node 
by mixing the embeddings of the neighboring nodes in the graph from the previous layer~\cite{bib16}.


\subsubsection{GraphSAGE}

The Graph Sample and Aggregate (GraphSAGE) algorithm was developed in~\cite{bib18}. 
Unlike GCNs, GraphSAGE randomly samples a fixed-size subset of node neighbors. 
This serves to limit the space and time complexity of the algorithm, irrespective of the graph 
structure and batch size. Similar to the convolution operation in CNNs, information relating to
the local neighborhood of a node is collected and used to compute the node embedding. 

At each iteration, the neighborhood of a node is initially sampled, and the information from the sampled 
nodes is ``aggregated'' into a single vector. The neighborhood of node~$v$
is denoted as~$\calN(v)$, where~$\calN(v)=\{u\,|\,(v,u)\in E\}$ for the graph~$G=(V,E)$. 
At layer~$k$, the aggregated information 
for node~$v$ is based on its neighborhood~$\calN(v)$; see~\cite{bib18} for more details.
The aggregation operation can be implemented as a mean, pooling, 
or LSTM function. We use the Pytorch \texttt{mean} as our aggregator. 

After applying the model trainable parameters 
and passing the result through a non-linear activation function such as ReLU, the layer~$k$ node~$v$ 
embedding is concatenated. The final representation (embedding) of node~$v$ is essentially 
the node embedding at the final layer. For node classification, this node embedding is passed 
through a sigmoid neuron or softmax layer.

\subsubsection{Graph Isomorphism Network}

The crucial difference between Graph Isomorphism Network (GIN)~\cite{bib18} and 
other GNNs is the message aggregation function. For GINs, this function is 
based on Multi-layer Perceptrons (MLP) at each layer.
Message aggregation methods are 
related to the Weisfeiler-Lehman algorithm mentioned above. 

\subsubsection{Simple Graph Convolution}

GCNs inherit considerable complexity from their deep learning lineage, 
which can be burdensome for less demanding applications. Simple Graph Convolution (SGC)~\cite{bib45} 
reduces the excess complexity of GCNs by repeatedly removing the nonlinearities between GCN layers and 
collapsing the resulting function into a single linear transformation. In contrast, a GCN transforms the feature vectors 
repeatedly throughout its layers and then applies a linear classifier on the final representation, 
SGC reduces the GCN procedure to a simple feature propagation step followed by standard logistic regression. 
%

\subsubsection{Jumping Knowledge Networks}

Similar to convolutional neural networks, GNN models of increasing depth can perform worse. While 
the message-passing mechanism helps us harness the information encapsulated in the graph structure, 
it may introduce some limitations if combined with GNN depth. 
In other words, our quest for a model that is more expressive and aware of the graph structure by adding 
more layers so that nodes can have a large receptive field can revert to a model that treats nodes all the same. 
This is called the over-smoothing problem~\cite{bib43}. To mitigate over-smoothing, we can apply the 
Jumping Knowledge~\cite{bib44} technique, which uses a concatenation layer. 

The key idea behind 
Jumping Knowledge is to select from all of those intermediate node representations and jump to the last 
layer, which serves to combine the intermediate node representations 
and to generate the final node representation. 
We apply layer aggregation concatenation to combine all intermediate node representations for
a linear transformation to compute the final node embeddings. The result then undergoes global pooling, 
with element-wise pooling over the final node embeddings. We apply Jumping Knowledge to 
the GCN, GraphSAGE, and GIN models, and we abbreviate these models
as JK-GCN, JK-GraphSAGE, and JK-GIN, respectively. 


\subsubsection{UnetGraph}

In~\cite{bib46}, U-Nets are used to correlate graph data to images by considering images as a special case of graphs, 
in which nodes lie on regular 2-D lattices. This structure enables us to use convolution and pooling operations on images. 
Therefore, node classification and embedding tasks have a natural correspondence with pixel-wise prediction tasks such 
as image segmentation~\cite{bib47}. In particular, both tasks aim to make predictions for each input unit, corresponding 
to a pixel on images or a node in graphs. In computer vision, pixel-wise prediction tasks have achieved major advances. 

Originally introduced in~\cite{bib48}, U-Nets~\cite{bib49}, which are based on an encoder-decoder architecture, 
are popular for 3-D image segmentation tasks. In addition to convolutions, pooling and up-sampling operations 
are essential building blocks in these architectures. However, extending these operations to graph data is 
challenging, because unlike grid-based data such as images and texts, nodes in graphs have no spatial locality 
and order information as required by regular pooling operations. To bridge this gap, the paper~\cite{bib46} introduces 
graph pooling (gPool) and unpooling (gUnpool) operations. We use the GraphUNet implementation of this architecture 
from the Torch Geometric library.
%

\subsubsection{Deep Graph Convolutional Neural Network}

Deep Graph Convolutional Neural Networks (DGCNN)~\cite{bib58} have three sequential stages. First,
graph convolution layers extract the local substructure of vertices features and define a consistent vertex ordering.
Second, a SortPooling layer sorts the vertex features under the previously defined order and unifies the input sizes.
Third, traditional convolutional and dense layers read the sorted graph representations and make predictions.

%

DGCNN has a number of advantages over other graph neural networks. For one, it directly accepts graph 
data as input without the need to first transform graphs into tensors, making end-to-end gradient-based training possible. 
Another potential advantage of DGCNN is that it enables learning from the global graph topology 
by sorting vertex features instead of summing them, which is supported by a novel \texttt{SortPooling} layer. 
DGCNN is the only sorting-based method that we consider. DGCNN was previously used in~\cite{bib58} 
for malware family classification using API call graphs.

\subsection{MalNet-Tiny Dataset}

As mentioned in Section~\ref{sect:datasets}, we use the MalNet-Tiny dataset,
 which consists of~5000 Function Call Graphs derived form Android \texttt{apk} files. 
 This is a balanced dataset, with each of the five families having~1000 samples. 
We highlight some basic features of these graphs in Table~\ref{tab:2}, where~$\sigma$ is the
standard deviation. From this data, we can clearly see that Downloader family should be 
relatively easy to distinguish from the other families.

\begin{table}[!htb]
\centering
\caption{Overview of graph features}\label{tab:2}
\adjustbox{scale=0.8}{
\begin{tabular}{c|c|c|c|c||c|c|c|c||c}\midrule\midrule
Malware & \multicolumn {4}{c||}{Number of vertices} & \multicolumn {4}{c||}{Number of edges} & Average \\
type & min & max & median & $\sigma$    & min & max & median & $\sigma$ & degree \\ \midrule
\texttt{AdDisplay}    & 122       & 4923   & 1248     & 1213.1         & 165       & 12,324     & 1895      & 2374.5         & 2.094 \\
\texttt{Adware}        & 211       & 4983   & 2317     & 1339.5         & 473       & 20,096     & 5381      & 3013.1         & 2.232 \\
\texttt{Benign}         & \zz\zz5 & 4994   & 1791      & 1510.5         & \zz\zz4 & 14,070      & 3591      & 2922.8         & 2.172 \\
\texttt{Downloader} & \zz40   & \zz117 & \zz\zz51 & \zz\zz\zz4.9 & \zz44    & \zz\zz143 & \zz\zz58 & \zz\zz\zz6.1 & 1.140 \\
\texttt{Trojan}         & \zz\zz9 & 4993    & \zz144   & 1313.5         & \zz\zz7  & 16,404      & \zz173   & 3275.2         & 2.156 \\ 
\midrule\midrule
\end{tabular}
}
\end{table}

\section{Experiments and Results}\label{chap:4}

In this section, we first consider results for non-GNN learning methods,
including MLP, which provide a baseline for comparison to GNN techniques. 
Then we consider extensive experiments with each of the GNN and JK models
discussed above.

\subsection{Results for Non-GNN Models}\label{sect:resutls-non-GNN}

Here, we provide experimental results for five methods, namely,
MLP, WL-Kernel, FEATHER, Slaq-LSD, and Slaq-VNGE. 
Each model requires a different set of hyperparameters, which we have determined via
a grid search in each case. For each model, the hyperparameters tested are 
specified in Table~\ref{tab:3}, where the selected values are given in boldface.

\begin{table}[!htb]
\centering
\caption{Hyperparameters and accuracy for non-GNN models}\label{tab:3}
\adjustbox{scale=0.85}{
\begin{tabular}{c|c|c|c}\midrule\midrule
Model & Hyperparameters & Tested values & Validation accuracy\\ \midrule
\multirow{3}{*}{MLP} & Number of layers & (\textbf{5}, 6) & \multirow{3}{*}{0.8054} \\ 
& Hidden dimensions & (\textbf{64}, 128) & \\ 
& Learning rate & (\textbf{0.001}, 0.0001) & \\ \midrule
WL-Kernel & Number of iterations & (5, \textbf{6}) & 0.7053 \\ \midrule
FEATHER & Order & (\textbf{2}, 5, 10) & 0.8488 \\  \midrule
\multirow{2}{*}{Slaq-LSD} & Number of vectors & (\textbf{10}, 15, 20) & \multirow{2}{*}{0.7799} \\  
& Number of steps & (\textbf{10}, 15, 20)  & \\ \midrule
\multirow{2}{*}{Slaq-VNGE} & Number of vectors & (10, 15, \textbf{20}) & \multirow{2}{*}{0.5499} \\  
& Number of steps & (10, 15, \textbf{20})  & \\ \midrule
\midrule
\end{tabular}
}
\end{table}

\subsubsection{MLP Results}

We use MLP as a benchmark for comparison to traditional deep learning models. 
In this case, the LDP features are used for classification. 
As can be observed from Table~\ref{tab:3}, our best result with accuracy of~$0.8054$ was obtained
using~$5$-layers trained for~$50$ epochs, with the dropout rate set to~$0.5$. 

\subsubsection{WL-Kernel Results}

We use node degree and page ranks and concatenated them into feature vectors for each node to 
generate features for the kernel-based method, WL-Kernel. The WL-Kernel method is used to compute a kernel matrix 
and then we apply Random Forest for classification, as discussed in Section~\ref{sect:WL}, above.
The model was tuned using the hyperparameters mentioned in Table~\ref{tab:3}, and the best accuracy 
attained was~0.7053. The accuracy and runtime metrics are summarized in Table~\ref{tab:B6}. 

\begin{table}[!htb]
\caption{Accuracy and runtime for WL-Kernel}\label{tab:B6}
\centering
\adjustbox{scale=0.85}{
\begin{tabular}{c|c|c|c|c}\midrule\midrule
\multirow{2}{*}{Iterations} & \multirow{2}{*}{Accuracy} & \multirow{2}{*}{Macro-F1} & Runtime & CPU \\ 
  & & & (in seconds) & cores \\ \midrule
\zz2 & 0.714 & 0.705 & 138.15 & 2 \\
\zz5 & 0.682 & 0.674 & 182.01 & 2 \\
  10 & 0.621 & 0.606 & 262.50 & 2 \\ \midrule\midrule
\end{tabular}
}
\end{table}

\subsubsection{FEATHER Results}

Using FEATHER, we perform a grid search over the order hyperparameter, 
which controls how much information is seen from higher-order neighborhoods. 
The accuracy and runtime metrics for these experiments
are given in Table~\ref{tab:B7}.

\begin{table}[!htb]
\caption{Accuracy and runtime for FEATHER model}\label{tab:B7}
\centering
\adjustbox{scale=0.85}{
\begin{tabular}{c|c|c|c}\midrule\midrule
\multirow{2}{*}{Order} & \multirow{2}{*}{Accuracy} & Runtime & CPU \\
  & & (in seconds) & cores \\ \midrule
4 & 0.839 & 672.78 & 2 \\
5 & 0.849 & 667.18  & 2 \\
6 & 0.847 & 680.34  & 2 \\ \midrule\midrule
\end{tabular}
}
\end{table}

\subsubsection{Slaq-LSD and Slaq-VNGE Results}

For Slaq-LSD and Slaq-VNGE, we perform a grid search over two hyperparameters, namely,
the number of random vectors and the number of Lanczos steps. 
The accuracy and runtime metrics for these models are given in Table~\ref{tab:B8}. 

%

\begin{table}[!htb]
\caption{Accuracy and runtime for Slaq-LSD and Slaq-VNGE}\label{tab:B8}
\centering
\adjustbox{scale=0.85}{
\begin{tabular}{c|c|c|c|c}\midrule\midrule
\multirow{2}{*}{Model} & \multirow{2}{*}{Vectors} & \multirow{2}{*}{Accuracy} & Runtime & CPU \\ 
  & & & (in seconds) & cores \\ \midrule
\multirow{3}{*}{Slaq-LSD} 
& 10 & 0.775 & 314.40 & 2 \\
& 15 & 0.776 & 352.35 & 2 \\
& 20 & 0.780 & 401.01 & 2 \\ \midrule
\multirow{3}{*}{Slaq-VNGE} 
& 10 & 0.549 & 300.85 & 2 \\
& 15 & 0.547 & 322.41  & 2 \\
& 20 & 0.544 & 361.35  & 2 \\ \midrule\midrule
\end{tabular}
}
\end{table}

\subsection{Results for GNN Architectures}

In this section, we experiment with each of the basic GNN models 
introduced in Section~\ref{sect:GNN-basic}. 
The input graph for each is a batched graph produced by \texttt{GraphDataLoader}. 
Each model is trained for~200 epochs. Each of the convolutional layers computes new node 
representations using a convolutional operator from the Pytorch Torch Geometric library,
and we use the Adam optimizer. 

Each model was tuned using the hyperparameters listed in Table~\ref{tab:4},
where the best hyperparameters are given in boldface.
Initial experiments were also conducted with a~$7$-layered GCN but virtually 
no improvement was observed while training time
increased considerably, and hence, the~$7$-layered model was abandoned. 

\begin{table}[!htb]
\centering
\caption{Hyperparameters and accuracy for GNN models}\label{tab:4}
\adjustbox{scale=0.85}{
\begin{tabular}{c|c|c|c}\midrule\midrule
Model & Hyperparameters & Tested values & Validation accuracy\\ \midrule
\multirow{3}{*}{GCN} & Number of layers & (5, \textbf{6}) & \multirow{3}{*}{0.9582} \\  
& Hidden dimensions & (64, \textbf{128}) & \\ 
& Learning Rate & (\textbf{0.001}, 0.0001) & \\ \midrule
\multirow{3}{*}{GraphSAGE} & Number of layers & (5, \textbf{6}) & \multirow{3}{*}{0.7913} \\  
& Hidden dimensions & (\textbf{64}, 128) & \\ 
& Learning rate & (\textbf{0.001}, 0.0001) & \\ \midrule
\multirow{3}{*}{GIN} & Number of layers & (5, \textbf{6}) & \multirow{3}{*}{0.9407} \\  
& Hidden dimensions & (\textbf{64}, 128)  & \\ 
& Learning rate & (\textbf{0.001}, 0.0001) & \\ \midrule
\multirow{3}{*}{SGC} & Number of layers & (\textbf{5}, 6) & \multirow{3}{*}{0.9079} \\  
& Hidden dimensions & (64, \textbf{128})  & \\ 
& Learning rate & (\textbf{0.001}, 0.0001) & \\ \midrule
\multirow{3}{*}{JK-GCN} & Number of layers & (5, \textbf{6}) & \multirow{3}{*}{0.8941} \\  
& Hidden dimensions & (64, \textbf{128}) & \\ 
& Learning rate & (\textbf{0.001}, 0.0001) & \\ \midrule
\multirow{3}{*}{JK-GraphSAGE} & Number of layers & (5, \textbf{6}) & \multirow{3}{*}{0.9291} \\  
& Hidden dimensions & (64, \textbf{128}) & \\ 
& Learning rate & (\textbf{0.001}, 0.0001) & \\ \midrule
\multirow{3}{*}{JK-GIN} & Number of layers & (5, \textbf{6}) & \multirow{3}{*}{0.9769} \\  
& Hidden dimensions & (64, \textbf{128})  & \\ 
& Learning rate & (\textbf{0.001}, 0.0001) & \\ \midrule
\multirow{3}{*}{UnetGraph} & Weight decay & (\textbf{0.001}, 0.0001) & \multirow{3}{*}{0.9581} \\  
& Hidden dimensions & (\textbf{64}, 128)  & \\ 
& Learning rate & (\textbf{0.001}, 0.0001) & \\ \midrule
\multirow{2}{*}{DGCNN} & Weight decay & (\textbf{0.001}, 0.0001) & \multirow{2}{*}{0.9218} \\ 
& Learning rate & (\textbf{0.001}, 0.0001) & \\ \midrule
\midrule
\end{tabular}
}
\end{table}

\subsubsection{GCN, GraphSAGE,  GIN, and SGC Results}

GCN uses the convolutional operator \texttt{GCNConv} 
and a global pooling layer to generate embeddings. 
GraphSAGE uses the convolutional operator \texttt{SAGEConv}, a batch normalization layer, 
and a global pooling layer to generate embeddings. 
GIN uses the convolutional operator \texttt{GINConv} and a global pooling layer to generate embeddings. 
SGC uses the convolutional operator \texttt{SGConv} and a global pooling layer to generate embeddings. 

From Table~\ref{tab:4}, we see that GCN performed the best of these models, with
an accuracy of~0.9582. In contrast, GraphSAGE performed the worst of these
four models, with an accuracy of~0.7913.

\subsubsection{Jumping Knowledge Model Results}

For JK-GCN, we use an extension of the PyTorch \texttt{torch.nn.Sequential} 
container to define a sequential GNN model. Since GNN operators take multiple input 
arguments, \texttt{torch\_geometric.nn.Sequential} expects both global input arguments 
and function header definitions of individual operators. If omitted, an intermediate 
module operates on the output of its preceding module. 
This allows us to create more sophisticated models, such as Jumping Knowledge models. 
The hyperparameters tested for each GNN model are listed in Table~\ref{tab:4},
where the best hyperparameters for each model are in boldface.

Similar to the GCN model, JK-GCN also uses \texttt{GCNConv} layers and implements 
them on a Jumping Knowledge model. The model was tuned using the hyperparameters 
mentioned in Table~\ref{tab:4}, and the best accuracy was~0.8941 


Our JK-GraphSAGE implementation uses \texttt{SAGEConv} layers and implements 
them on a Jumping Knowledge model. The model was tuned using the hyperparameters mentioned in 
Table~\ref{tab:4}, and the best accuracy obtained was~0.9291. 


Our JK-GIN model uses \texttt{GINConv} layers and implements them 
on a Jumping Knowledge model. The best accuracy in this case was~0.9769. 

\subsubsection{UnetGraph and DGCNN Results}

We create this model using Pytorch \texttt{UnetGraph}, with a depth of~$3$ and pooling ratios 
set to~$1$ and~$0.5$. We also set the dropout rate to~$0.5$. The model was tuned using the hyperparameters 
in Table~\ref{tab:4}, and we trained for~100 epochs instead of the~200 used for previous models 
due to over-fitting at larger numbers of epochs.
The best accuracy achieved by our UnetGraph models was~0.9581. 


Our DGCNN model consists of~4 layers of \texttt{GCNConv}, 
a pair of~1-D convolutional layers, and a max pooling layer. 
Two linear classifiers are used to concatenate these layers, with a dropout rate of~$0.5$. 
The model was tuned using the hyperparameters mentioned in Table~\ref{tab:4}, 
and the best accuracy was found to be~0.9218. 
With additional hyperparameter tuning, 
it is likely that this model can be significantly improved, 
since we only varied the learning rate and weight decay. 

\subsection{Discussion}\label{chap:5}

In this section, we provide additional context for the results in the previous section. 
Specifically, we summarize the accuracies, classwise accuracies, and training efficiencies
for each of the various models. We also provide UMAP embeddings, which allow
us to visualize the models.

For clarity, in each case we split the analysis into two categories, namely, 
graph learning methods as introduced in Section~\ref{sect:non-GNN-learning} 
(that is, MLP, WL-Kernel, FEATHER, Slaq-VNGE, and Slaq-LSD)
and the GNN models introduced in Section~\ref{sect:GNN-basic} 
(i.e., GCN, GraphSAGE, GIN, SGC, JK-GCN, JK-GraphSAGE, JK-GIN, UnetGraph, 
and DGCNN). Here, we refer to the former as non-GNN models,
while the latter are GNN models.

\subsubsection{Accuracy Comparison}

\begin{figure}[!htb]
\centering
\begin{tabular}{cc}
    \begin{tikzpicture}[scale=0.5, every node/.style={scale=1.0}]
    \begin{axis}[
        width  = 0.825*\textwidth,
        height = 9.5cm,
        ymin=0.0,ymax=1.2,
        ytick={0,0.2,0.4,0.6,0.8,1.0},
        major x tick style = transparent,
        ybar=5*\pgflinewidth,
        bar width=42.0pt,
        ylabel = {Accuracy},
        symbolic x coords={
MLP,
WL-Kernel,
FEATHER,
Slaq-LSD,
Slaq-VGNE
},
        xtick=data,
	y tick label style={
    		/pgf/number format/.cd,
   		fixed,
   		fixed zerofill,
    		precision=2},
        x tick label style={
        		rotate=60,
		font=\tt,
		anchor=north east,
		inner sep=0mm
		},
        nodes near coords,
        every node near coord/.append style={rotate=90, 
        								   anchor=west,
								   /pgf/number format/.cd,
								   	fixed zerofill,
									precision=4
								   },
        enlarge x limits=0.15,
        legend cell align=left,
        legend pos=south east,
    ]
\addplot[fill=blue,opacity=1.00]
coordinates {
(MLP,0.8054)
(WL-Kernel,0.7053)
(FEATHER,0.8488)
(Slaq-LSD,0.7799)
(Slaq-VGNE,0.5499)
};
\end{axis}
\end{tikzpicture}
    &
    \raisebox{-7.5pt}{\begin{tikzpicture}[scale=0.5, every node/.style={scale=1.0}]
    \begin{axis}[
        width  = 0.825*\textwidth,
        height = 9.5cm,
        ymin=0.0,ymax=1.2,
        ytick={0,0.2,0.4,0.6,0.8,1.0},
        major x tick style = transparent,
        ybar=5*\pgflinewidth,
        bar width=25.0pt,
        ylabel = {Accuracy},
        symbolic x coords={
GCN,
GraphSAGE,
GIN,
SGC,
JK-GCN,
JK-GraphSAGE,
JK-GIN,
UnetGraph,
DGCNN
},
        xtick=data,
	y tick label style={
    		/pgf/number format/.cd,
   		fixed,
   		fixed zerofill,
    		precision=2},
        x tick label style={
        		rotate=60,
		font=\tt,
		anchor=north east,
		inner sep=0mm
		},
        nodes near coords,
        every node near coord/.append style={rotate=90, 
        								   anchor=west,
								   /pgf/number format/.cd,
								   	fixed zerofill,
									precision=4
								   },
        enlarge x limits=0.1,
        legend cell align=left,
        legend pos=south east,
    ]
\addplot[fill=red,opacity=1.00] 
coordinates {
(GCN,0.9582)
(GraphSAGE,0.7913)
(GIN,0.94070)
(SGC,0.9097)
(JK-GCN,0.8941)
(JK-GraphSAGE,0.9291)
(JK-GIN,0.9769)
(UnetGraph,0.9581)
(DGCNN,0.9218)
};
\end{axis}
\end{tikzpicture}}
    \\
    \adjustbox{scale=0.85}{(a) Non-GNN models}
    &
    \adjustbox{scale=0.85}{(b) GNN models}
\end{tabular}
\caption{Accuracy comparison}\label{fig:22-23}
\end{figure}
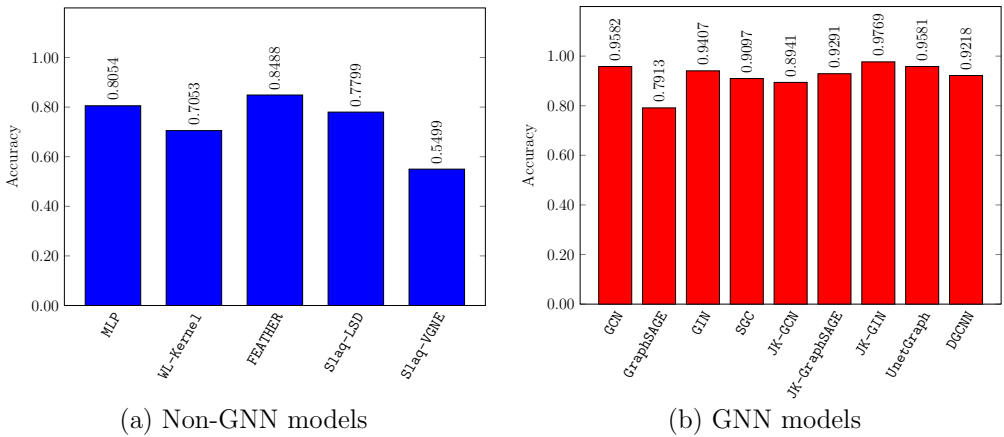

In Figure~\ref{fig:22-23}(a) we compare the accuracies of our non-GNN models, while
Figure~\ref{fig:22-23}(b) is an analogous comparison of the GNN-based models that we tested. 
Most of the GNN-based models far outperform the best of the non-GNN models. 
We observe that, overall, JK-GIN has the highest accuracy at~0.9769, 
with UnetGraph having an accuracy of~0.9581, while GIN (0.9407), 
JK-GraphSAGE (0.9291), DGCNN (0.9218) and SGC (0.9079) also outperform
the best of the non-GNN models. While GCN has a high accuracy of~0.9582, we believe 
it may be due to over-smoothing. 
As mentioned above, we believe that further tuning of the 
hyperparameters might yield significant improvement for DGCNN. 

As can be seen in Figure~\ref{fig:22-23}(b), among the non-GNN-based models, 
FEATHER was the best at~0.8488, with MLP being second best at~0.8054. 
These results are not unexpected, as these models were shown to work well
(along with GIN) in the paper~\cite{bib21}.
The bottom line is that our best GNN-based models outperform all models that 
were applied to the MalNet-Tiny dataset in the papers~\cite{bib21,bib57};
see also Table~\ref{tab:B9}, above.

\subsubsection{Classwise Comparison}

To further analyze our results, we generate heat maps of the 
accuracy of each class for 
every model corresponding to its best parameters. 
In Figure~\ref{fig:24-25}(a), we observe that our worst-performing 
GNN-based models, namely, GraphSAGE and JK-GCN, underperform
primarily because they badly fail on the benign class. 
We also note that all of the GNN-based models
classify the Downloader class with ease.

Classwise comparison of our non-GNN models is given in Figure~\ref{fig:24-25}(b).
Qualitatively, the results for these models are similar to those of the GNN-based
models, with \texttt{Downloader} being the easiest to classify,
and \texttt{Benign} being the most challenging. However, as noted above,
the non-GNN models
perform poorly, in comparison to the best of our GNN-based models.

\begin{figure}[!htb]
\centering
\begin{tabular}{c}
\raisebox{8.25pt}{
\begin{tikzpicture}[scale=0.45]
    \begin{axis}[
        width=9.5cm,
        height=10cm,
	colormap={redblack}{color=(black) color=(red)},
        xticklabels={
\texttt{FEATHER},
\texttt{Slaq-LSD},
\texttt{Slaq-VNGE},
\texttt{WL-Kernel},
\texttt{MLP}
        },
        xtick={0,...,4},
        xtick style={draw=none},
	xticklabel style={anchor=east,rotate=60,yshift=-5pt,font=\tt\large},
        yticklabels={
\texttt{AdDisplay},
\texttt{Adware},
\texttt{Benign},
\texttt{Downloader},
\texttt{Trojan},
\texttt{Weighted Average}
        },
        ytick={0,...,5},
        ytick style={draw=none},
        enlargelimits=false,
        yticklabel style={font=\tt\large},
        colorbar,
        colorbar style={
            ytick={0.5,0.6,0.7,0.8,0.9,1.0},
            yticklabels={0.5,0.6,0.7,0.8,0.9,1.0},
            yticklabel={\pgfmathprintnumber\tick},
            yticklabel style={
            		scale=1.2,
            		/pgf/number format/fixed,
			/pgf/number format/precision=1}
        },
        point meta min=0.5,
        point meta max=1.0,
        nodes near coords={\pgfmathprintnumber\pgfplotspointmeta},
        nodes near coords black white/.style={
            small value/.style={
                yshift=-7pt,
                text=black,
                /pgf/number format/fixed,
                /pgf/number format/precision=2,
                /pgf/number format/zerofill=true,
                scale=1.2,
            },
            large value/.style={
                yshift=-7pt,
                text=white,
                /pgf/number format/fixed,
                /pgf/number format/precision=2,
                /pgf/number format/zerofill=true,
                scale=1.2,
            },
            every node near coord/.style={
                check for zero/.code={
                    \pgfmathfloatifflags{\pgfplotspointmeta}{0}{
                        \pgfkeys{/tikz/coordinate}
                    }{
                        \begingroup
                        \pgfkeys{/pgf/fpu}
                        \pgfmathparse{\pgfplotspointmeta<#1}
                        \global\let\result=\pgfmathresult
                        \endgroup
                        %
                        %
                        \pgfmathfloatcreate{1}{1.0}{0}
                        \let\ONE=\pgfmathresult
                        \ifx\result\ONE
                            \pgfkeysalso{/pgfplots/small value}
                        \else
                            \pgfkeysalso{/pgfplots/large value}
                        \fi
                    }
                },
                check for zero,
            },
        },
        nodes near coords black white=0.1,
    ]
        \addplot[
            matrix plot,
            mesh/cols=5,
            point meta=explicit,draw=gray
        ] table [meta=C] {
            x y C
0 0 0.83
1 0 0.78
2 0 0.50
3 0 0.66
4 0 0.72
0 1 0.83
1 1 0.73
2 1 0.44
3 1 0.70
4 1 0.80
0 2 0.72
1 2 0.62
2 2 0.27
3 2 0.47
4 2 0.63
0 3 0.99
1 3 0.98
2 3 0.93
3 3 0.97
4 3 0.99
0 4 0.86
1 4 0.79
2 4 0.61
3 4 0.73
4 4 0.99
0 5 0.85
1 5 0.78
2 5 0.55
3 5 0.71
4 5 0.81
         };
    \end{axis}
\end{tikzpicture}}
\\[-2.25ex]
\adjustbox{scale=0.85}{(a) Non-GNN models}
\\
\\[-2ex]
\begin{tikzpicture}[scale=0.45]
    \begin{axis}[
        width=15cm,
        height=10cm,
	colormap={redblack}{color=(black) color=(red)},
        xticklabels={
\texttt{GCN},
\texttt{GIN},
\texttt{GraphSAGE},
\texttt{SGC},
\texttt{JK-GCN},
\texttt{JK-GIN},
\texttt{JK-GraphSAGE},
\texttt{UnetGraph},
\texttt{DGCNN}
        },
        xtick={0,...,8},
        xtick style={draw=none},
	xticklabel style={anchor=east,rotate=60,yshift=-5pt,font=\tt\large},
        yticklabels={
\texttt{AdDisplay},
\texttt{Adware},
\texttt{Benign},
\texttt{Downloader},
\texttt{Trojan},
\texttt{Weighted Average}
        },
        ytick={0,...,5},
        ytick style={draw=none},
        enlargelimits=false,
        yticklabel style={font=\tt\large},
        colorbar,
        colorbar style={
            ytick={0.5,0.6,0.7,0.8,0.9,1.0},
            yticklabels={0.5,0.6,0.7,0.8,0.9,1.0},
            yticklabel={\pgfmathprintnumber\tick},
            yticklabel style={
            		scale=1.2,
            		/pgf/number format/fixed,
			/pgf/number format/precision=1}
        },
        point meta min=0.5,
        point meta max=1.0,
        nodes near coords={\pgfmathprintnumber\pgfplotspointmeta},
        nodes near coords black white/.style={
            small value/.style={
                yshift=-7pt,
                text=black,
                /pgf/number format/fixed,
                /pgf/number format/precision=2,
                /pgf/number format/zerofill=true,
                scale=1.2,
            },
            large value/.style={
                yshift=-7pt,
                text=white,
                /pgf/number format/fixed,
                /pgf/number format/precision=2,
                /pgf/number format/zerofill=true,
                scale=1.2,
            },
            every node near coord/.style={
                check for zero/.code={
                    \pgfmathfloatifflags{\pgfplotspointmeta}{0}{
                        \pgfkeys{/tikz/coordinate}
                    }{
                        \begingroup
                        \pgfkeys{/pgf/fpu}
                        \pgfmathparse{\pgfplotspointmeta<#1}
                        \global\let\result=\pgfmathresult
                        \endgroup
                        %
                        %
                        \pgfmathfloatcreate{1}{1.0}{0}
                        \let\ONE=\pgfmathresult
                        \ifx\result\ONE
                            \pgfkeysalso{/pgfplots/small value}
                        \else
                            \pgfkeysalso{/pgfplots/large value}
                        \fi
                    }
                },
                check for zero,
            },
        },
        nodes near coords black white=0.5,
    ]
        \addplot[
            matrix plot,
            mesh/cols=9,
            point meta=explicit,draw=gray
        ] table [meta=C] {
            x y C
0 0 0.99
1 0 0.97
2 0 0.89
3 0 0.91
4 0 0.77
5 0 0.99
6 0 0.94
7 0 0.99
8 0 0.91
0 1 0.95
1 1 0.94
2 1 0.72
3 1 0.94
4 1 0.90
5 1 0.95
6 1 0.94
7 1 0.94
8 1 0.93
0 2 0.92
1 2 0.89
2 2 0.57
3 2 0.81
4 2 0.57
5 2 0.93
6 2 0.86
7 2 0.93
8 2 0.84
0 3 0.99
1 3 0.98
2 3 0.97
3 3 0.98
4 3 0.99
5 3 0.99
6 3 0.99
7 3 0.99
8 3 0.99
0 4 0.94
1 4 0.92
2 4 0.81
3 4 0.91
4 4 0.87
5 4 0.97
6 4 0.91
7 4 0.94
8 4 0.94
0 5 0.96
1 5 0.94
2 5 0.79
3 5 0.91
4 5 0.82
5 5 0.97
6 5 0.93
7 5 0.96
8 5 0.92
         };
    \end{axis}
\end{tikzpicture}
\\[-1.25ex]
\adjustbox{scale=0.85}{(b) GNN-based models}
\end{tabular}
\caption{Classwise accuracies}\label{fig:24-25}
\end{figure}

Overall, we see that \texttt{Downloader} is the easiest class to classify,
while the \texttt{Benign} class is the most difficult.  Even our worst model, 
Slaq-VNGE, identifies \texttt{Downloader} class with a~0.93 accuracy, 
while our best classifiers attain a similar accuracy for the \texttt{Benign} class.
These results are not surprising in light of our initial analysis 
of the node and edge features, as given in Table~\ref{tab:2}, above.

We also provide confusion matrices for each of the non-GNN
models and for the GNN models in Figures~\ref{fig:conf_non} and~\ref{fig:conf_GNN},
which can be found in the Appendix.
From these confusion matrices, we observe that the \texttt{Benign} class is most often
misclassified as belonging to the \texttt{Adware} or \texttt{AdDisplay} class. Interestingly,
we do not find a large percentage of misclassifications of \texttt{Benign}
as \texttt{Trojan} (or vice-versa) which, intuitively, would seem to be
difficult to distinguish, as Trojans are designed to mimic benign applications.

\subsubsection{Runtime Comparison}

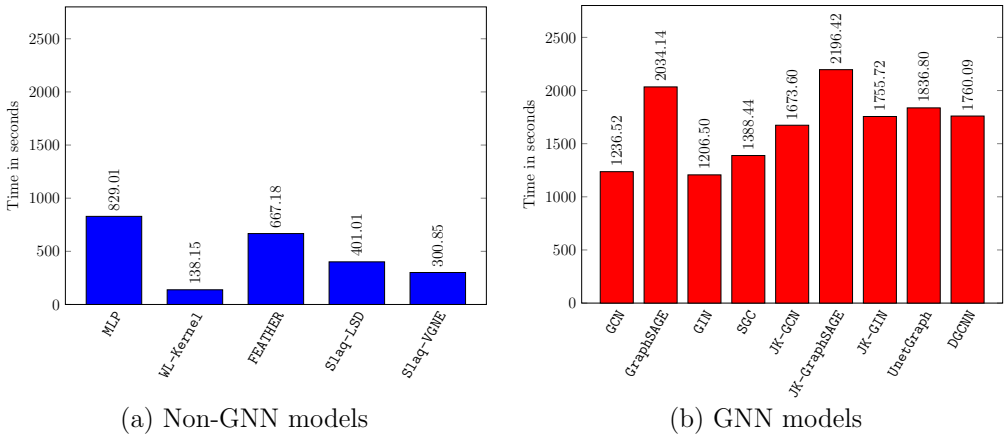
\begin{figure}[!htb]
\centering
\begin{tabular}{cc}
    \begin{tikzpicture}[scale=0.5, every node/.style={scale=1.0}]
    \begin{axis}[
        width  = 0.825*\textwidth,
        height = 9.5cm,
        ymin=0,ymax=2800,
        ytick={0,500,1000,1500,2000,2500},
        major x tick style = transparent,
        ybar=5*\pgflinewidth,
        bar width=42.0pt,
        ylabel = {Time in seconds},
        symbolic x coords={
MLP,
WL-Kernel,
FEATHER,
Slaq-LSD,
Slaq-VGNE
},
        xtick=data,
	y tick label style={
    		/pgf/number format/.cd,
   		fixed,
   		fixed zerofill,
    		precision=0,
		1000 sep={}},
        x tick label style={
        		rotate=60,
		font=\tt,
		anchor=north east,
		inner sep=0mm
		},
        nodes near coords,
        every node near coord/.append style={rotate=90, 
        								   anchor=west,
								   /pgf/number format/.cd,
								   	fixed zerofill,
									precision=2
								   },
        enlarge x limits=0.15,
        legend cell align=left,
        legend pos=south east,
    ]
\addplot[fill=blue,opacity=1.00]
coordinates {
(MLP,829.01)
(WL-Kernel,138.15)
(FEATHER,667.18)
(Slaq-LSD,401.01)
(Slaq-VGNE,300.85)
};
\end{axis}
\end{tikzpicture}
    &
    \raisebox{-7.5pt}{\begin{tikzpicture}[scale=0.5, every node/.style={scale=1.0}]
    \begin{axis}[
        width  = 0.825*\textwidth,
        height = 9.5cm,
        ymin=0,ymax=2800,
        ytick={0,500,1000,1500,2000,2500},
        major x tick style = transparent,
        ybar=5*\pgflinewidth,
        bar width=25.0pt,
        ylabel = {Time in seconds},
        symbolic x coords={
GCN,
GraphSAGE,
GIN,
SGC,
JK-GCN,
JK-GraphSAGE,
JK-GIN,
UnetGraph,
DGCNN
},
        xtick=data,
	y tick label style={
    		/pgf/number format/.cd,
   		fixed,
   		fixed zerofill,
    		precision=0,
		1000 sep={}},
        x tick label style={
        		rotate=60,
		font=\tt,
		anchor=north east,
		inner sep=0mm
		},
        nodes near coords,
        every node near coord/.append style={rotate=90, 
        								   anchor=west,
								   /pgf/number format/.cd,
								   	fixed zerofill,
									precision=2,
									1000 sep={}
								   },
        enlarge x limits=0.1,
        legend cell align=left,
        legend pos=south east,
    ]
\addplot[fill=red,opacity=1.00]
coordinates {
(GCN,1236.52)
(GraphSAGE,2034.14)
(GIN,1206.50)
(SGC,1388.44)
(JK-GCN,1673.60)
(JK-GraphSAGE,2196.42)
(JK-GIN,1755.72)
(UnetGraph,1836.80)
(DGCNN,1760.09)
};
\end{axis}
\end{tikzpicture}}
    \\
    \adjustbox{scale=0.85}{(a) Non-GNN models}
    &
    \adjustbox{scale=0.85}{(b) GNN models}
\end{tabular}
\caption{Runtime comparison}\label{fig:26-27}
\end{figure}

We define the runtime to be the total time taken to train, test, and validate a model 
using its corresponding best set of hyperparameters.
The runtime results for our GNN-based models are given in Figure~\ref{fig:26-27}(a). 
We observe that both GraphSAGE and JK-GraphSAGE run for longer durations than other models. 
The convolution operator, \texttt{SAGEConv}, is the most complex operator used in any of the GNNs, 
so this result is expected. Among the GNN-based models, GIN and GCN models require the least time 
for classification, as they use the least complex convolutional operators.

The runtime results for our non-GNN models are given in Figure~\ref{fig:26-27}(b). 
These models require significantly less time than GNN-based models, with
the longest runtime among this collection of models being only about two-thirds
that of the shortest runtime among the GNN-based models. Note that
the best performing non-GNN models, FEATHER and MLP, have the longest runtimes.

\subsubsection{UMAP Embeddings for GNN Models}

Finally, we provide a visualization of the graph embeddings for nine of our models. 
In each case, we apply the UMAP dimensionality reduction technique~\cite{bib56}
to the dense layer. These results are given in Figure~\ref{fig:28} in the Appendix, where 
we have labeled the classes as
$$ 
  (\mbox{\texttt{AdDisplay}},
  \mbox{\texttt{Adware}}, 
  \mbox{\texttt{Benign}}, 
  \mbox{\texttt{Downloader}},
  \mbox{\texttt{Trojan}})=(0,1,2,3,4).
$$

In Figure~\ref{fig:28} we observe that there is generally the most class separability between 
\texttt{Downloader}, and other classes, which is supported by our observation that \texttt{Downloader} 
is the easiest to classify. We also note that several of the models provide good class separability. In particular,  
the jumping knowledge models, JK-GCN, JK-GraphSAGE, and JK-GIN, which appear in
Figures~\ref{fig:28}(e), (f), and~(g), respectively, are among the best from this perspective. 

Overall, these UMAP embeddings support some of our major observations. They also 
show that our better models tend to form numerous small clusters, 
which likely correspond to identifiable sub-classes within each class.
This would be an interesting topic for further research.

\section{Conclusion and Future Work}\label{chap:6}

In this extensive study, we classified five different types of Android samples from 
the the MalNet-Tiny dataset graph-based learning methods.
We considered the graph representation schemes FEATHER, WL-Kernel, 
Slaq-VNGE, and Slaq-LSD, and applied a Random Forest classifier to each. 
We found that FEATHER works best among these methods. 

We then used the Local Degree Profile (LDP) to encode node features
and trained a wide variety of GNN-based models on these features.
Specifically, we experimented with GCN, GrahSAGE, GIN, SGC, JK-GCN, JK-GraphSAGE, JK-GIN, 
UnetGraph, and DGCNN models. We experimented with hyperparameters
and found that these GNN-based models generally performed well, 
with JK-GIN, JK-GraphSAGE, UnetGraph, and DGCNN giving us the best results. 
We provided extensive analysis of our results, including classwise accuracies,
confusion matrices, UMAP embeddings, and runtime comparisons.
Our best models exceeded the results obtained in comparable previous work.

In future work, we intend to extend this promising research to a larger subset of the MalNet dataset.
A more diverse dataset, with samples drawn from a larger number of families, would provide
additional scope for tuning hyperparameters and analyzing the resulting models,
so as to better understand their relative strengths and weaknesses. 

It would be interesting to explore other graph classification architectures, including 
\hbox{DeeperGCN}~\cite{bib51}, \hbox{EdgePool}~\cite{bib50}, and Kernel Graph-based CNN (KG-CNN)~\cite{bib54}. 
It would also be worth exploring edge-based graph neural approaches such as \hbox{E-GraphSAGE}~\cite{bib52},
which would enable us to include network flows as features. 

Another area of future research would be to analyze the effectiveness of GNN-based models
for zero-day malware detection. In a similar vein, it would also be interesting to explore malware 
GNN models from the perspective of explainable AI (XAI), that is, we would like to better understand how the models
are actually making decisions. For this explainability problem, we could use \hbox{GNNExplainer}~\cite{bib53},
which is designed to derive insights from the hidden layers of GNNs, and \hbox{SubgraphX}~\cite{bib55}.

Finally, it is worth noting that the models considered in this paper are based solely on graph structure. 
Combining such graph-based models with more traditional features might result in models that are
stronger than either the GNN-based models considered here, or learning models that are based
only on more traditional features.

\section{Declarations}

The authors have no relevant financial or non-financial interests to disclose.

\bibliographystyle{plain}
\bibliography{references.bib}

\appendix
\renewcommand\thefigure{A.\arabic{figure}}    
\section*{Appendix}
\setcounter{figure}{0}    

\begin{figure}[!htb]
\centering
\advance\tabcolsep by -6pt
\begin{tabular}{ccc}
\includegraphics[width=0.25\linewidth]{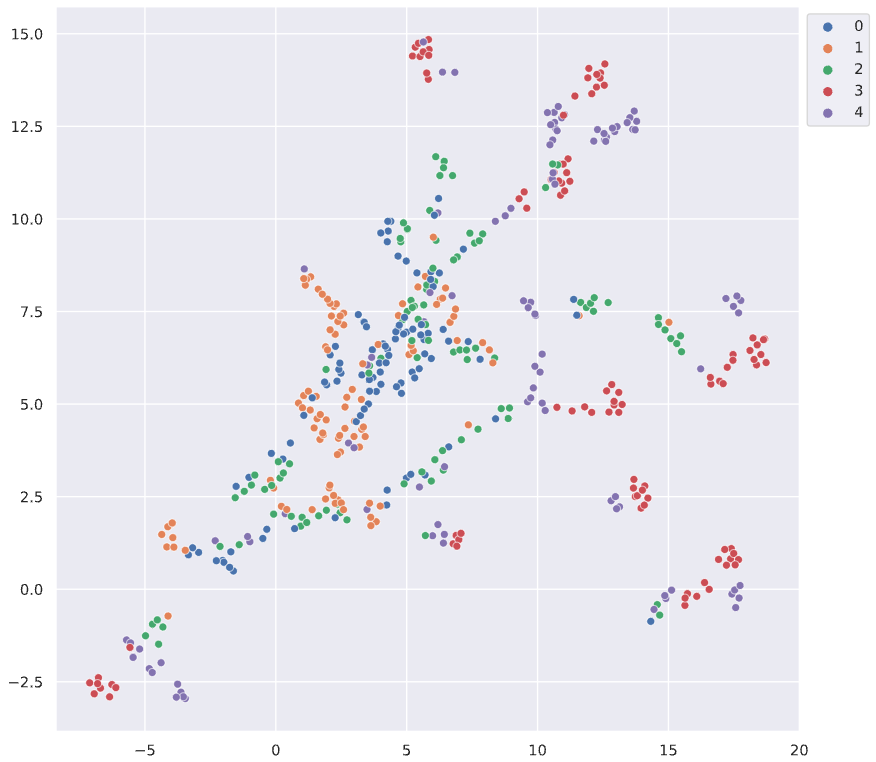}
&
\includegraphics[width=0.245\linewidth]{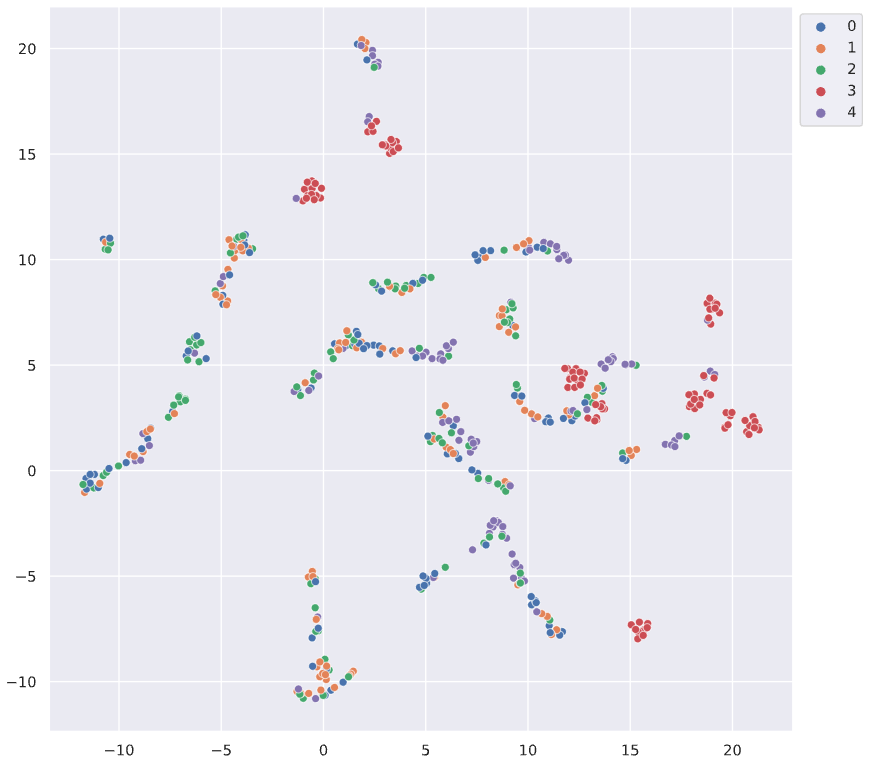}
&
\includegraphics[width=0.25\linewidth]{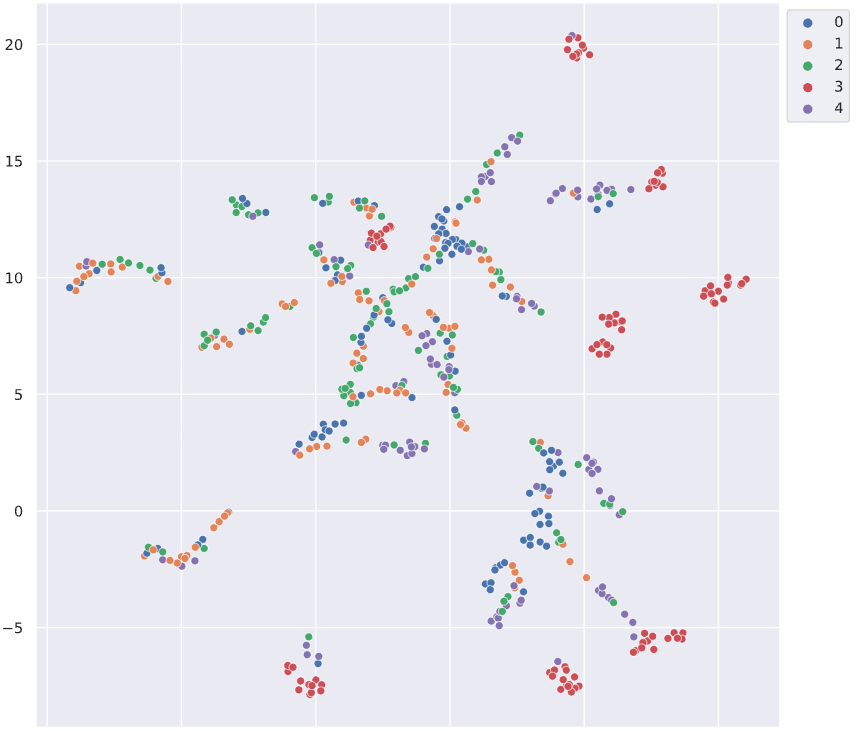}
\\
\adjustbox{scale=0.85}{(a) GCN}
&
\adjustbox{scale=0.85}{(b) GraphSAGE}
&
\adjustbox{scale=0.85}{(c) GIN}
\\[-1ex]
\\
\includegraphics[width=0.25\linewidth]{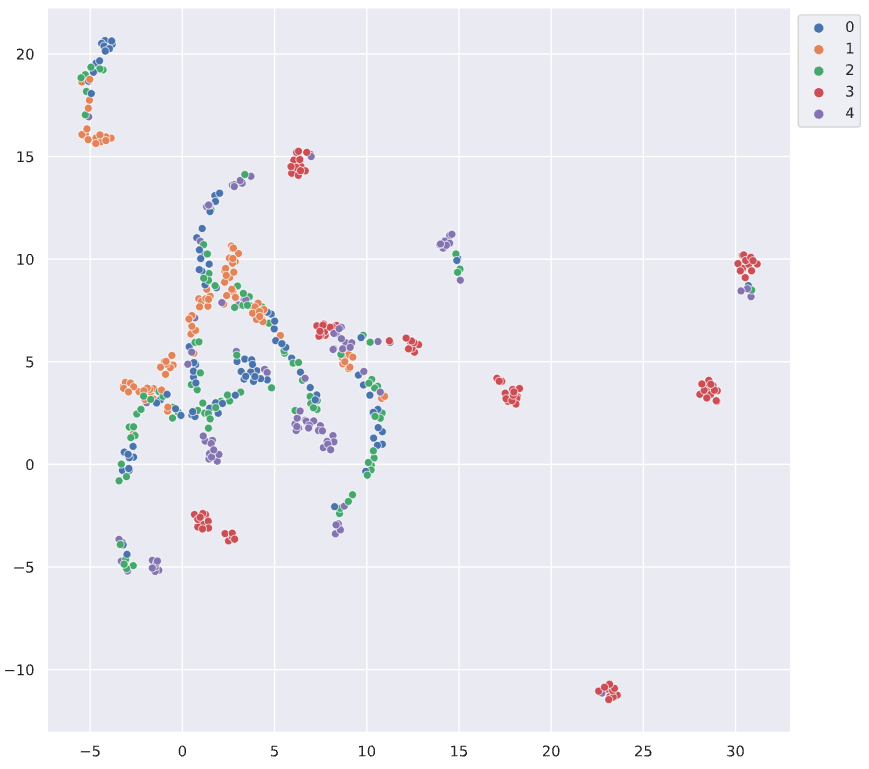}
&
\includegraphics[width=0.25\linewidth]{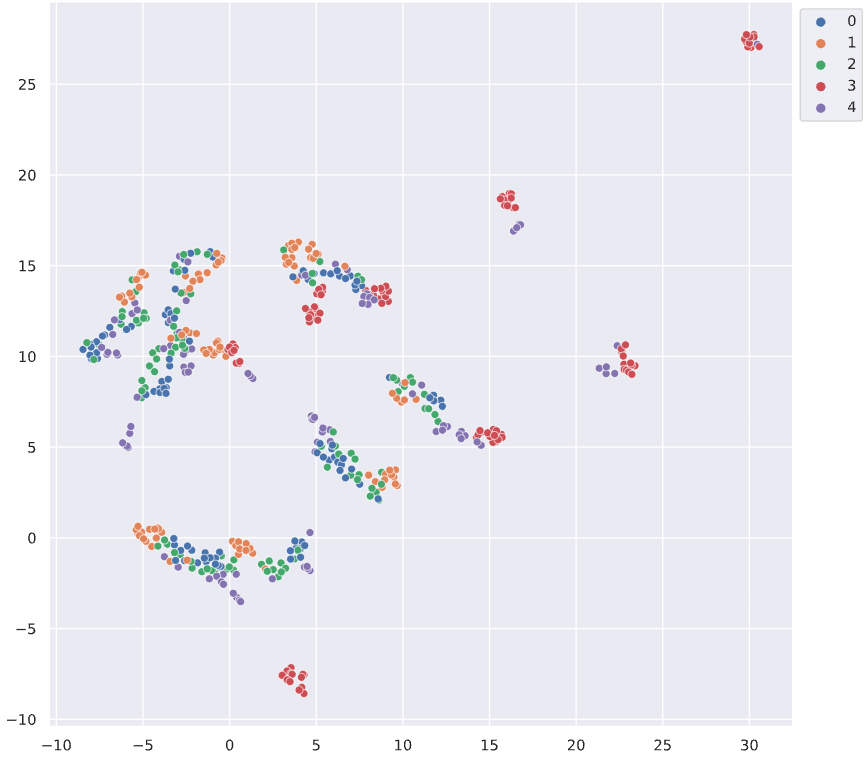}
&
\includegraphics[width=0.25\linewidth]{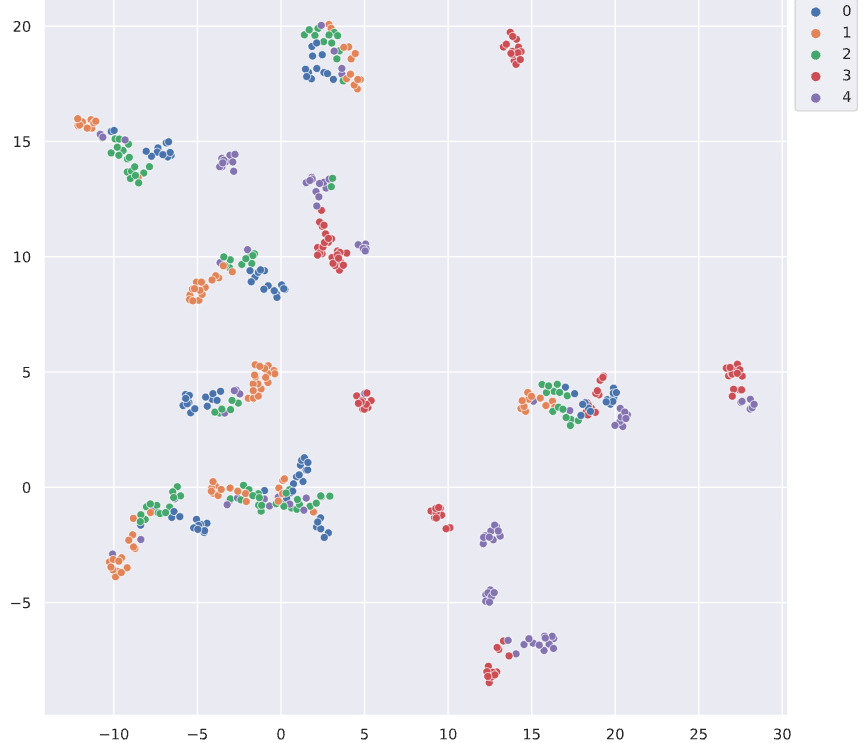}
\\
\adjustbox{scale=0.85}{(d) SGC}
&
\adjustbox{scale=0.85}{(e) JK-GCN}
&
\adjustbox{scale=0.85}{(f) JK-GraphSAGE}
\\[-3.0ex]
\\
\raisebox{10pt}{\includegraphics[width=0.245\linewidth]{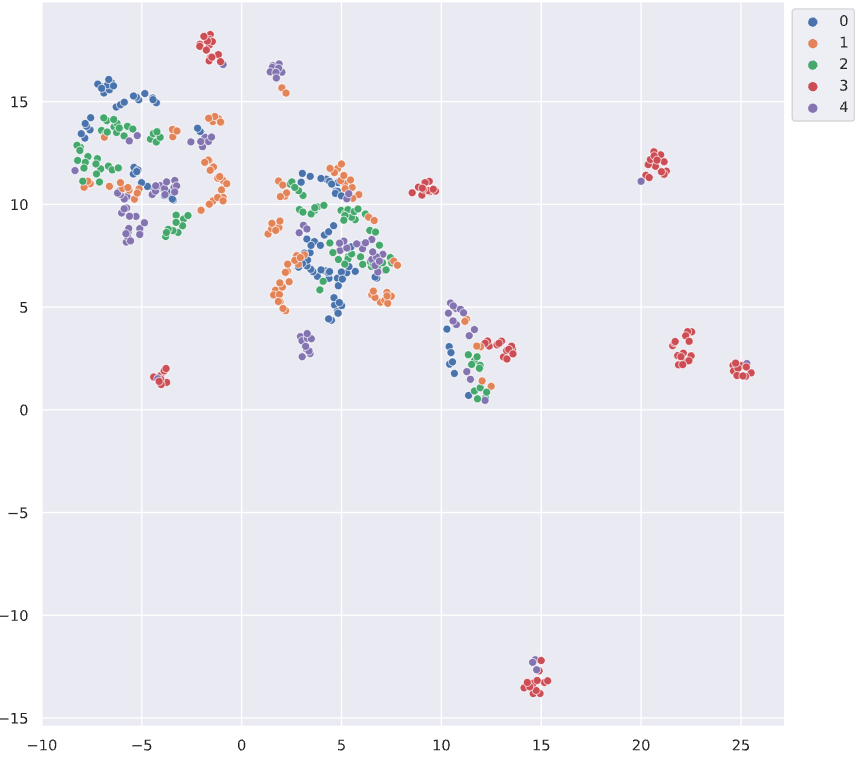}}
&
\includegraphics[width=0.275\linewidth]{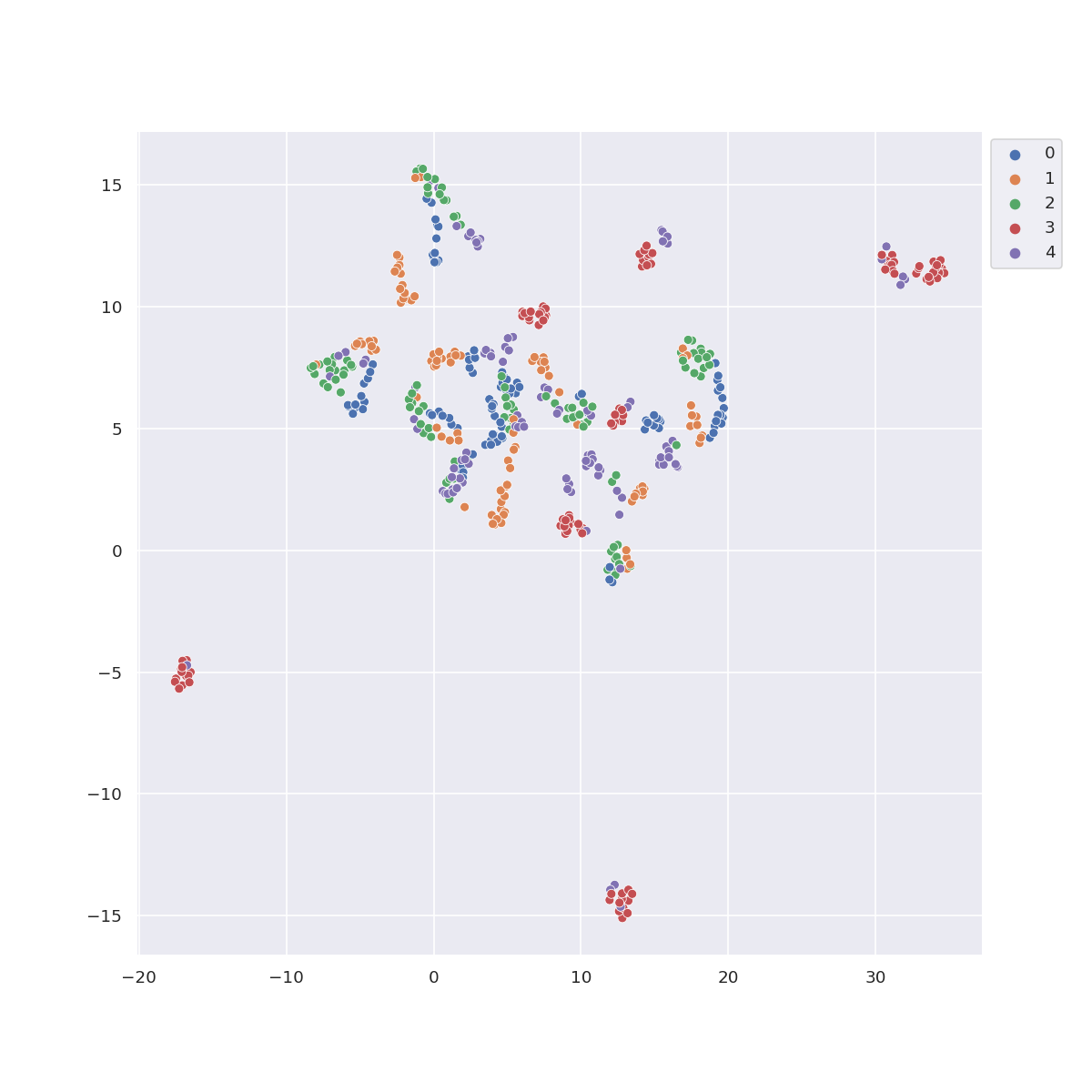}
&
\includegraphics[width=0.275\linewidth]{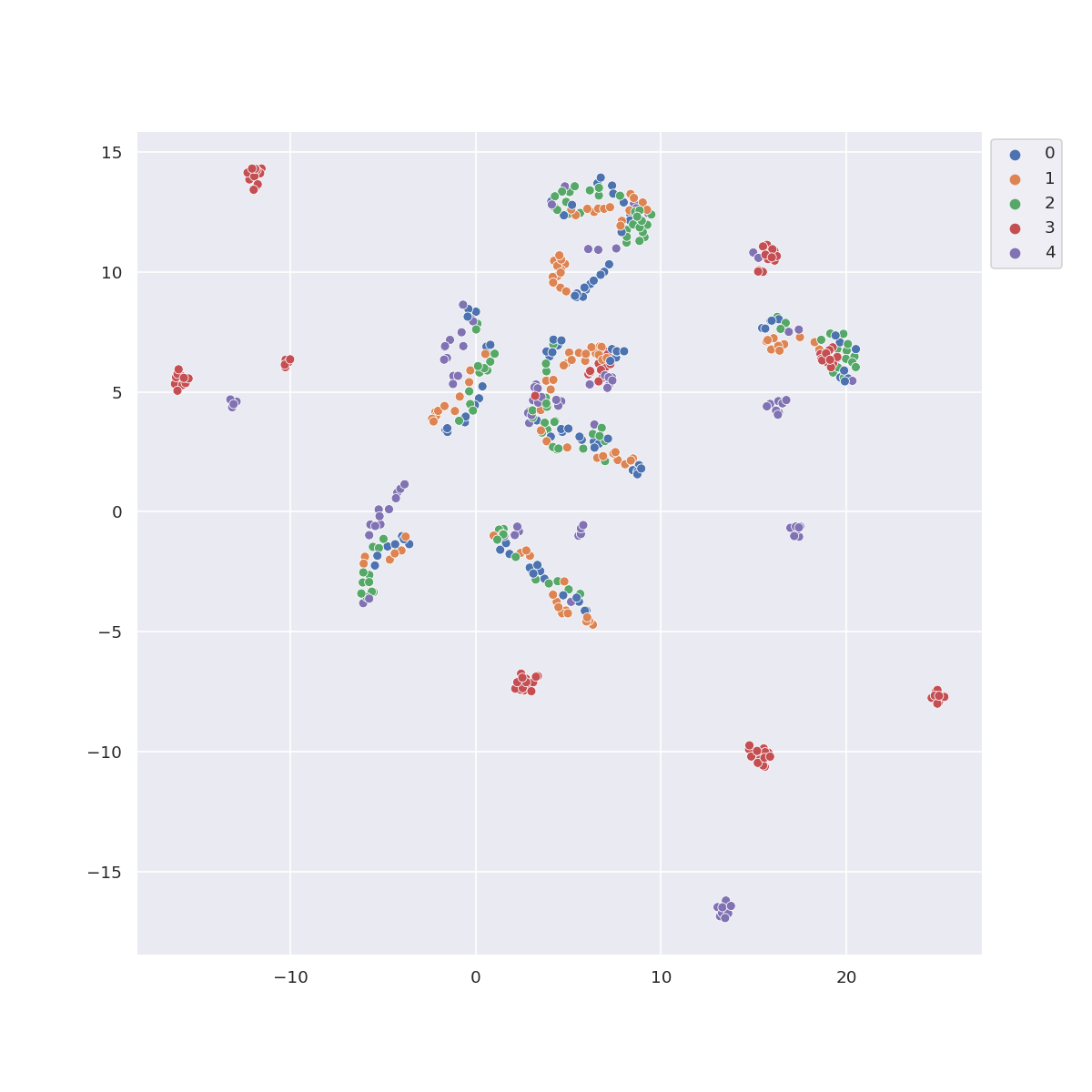}
\\
\adjustbox{scale=0.85}{(g) JK-GIN}
&
\adjustbox{scale=0.85}{(h) UnetGraph}
&
\adjustbox{scale=0.85}{(i) DGCNN}
\end{tabular}
\caption{UMAP embeddings}\label{fig:28}
\end{figure}

This appendix contains confusion matrices and UMAP embeddings for models considered in this paper.
In Figure~\ref{fig:conf_non}, we have provided confusion matrices for all five of our non-GNN models,
while Figure~\ref{fig:conf_GNN} has confusion matrices for all nine of our GNN-based models.
The UMAP embeddings for the nine GNN-based models appear in Figure~\ref{fig:28}.

\begin{figure}[!htb]
\centering
\begin{tabular}{cc}
\begin{tikzpicture}[scale=0.4]
    \begin{axis}[
        width=10cm,
        height=10cm,
	colormap={bluewhite}{color=(white) rgb255=(100,149,237)},
        xticklabels={
\texttt{AdDisplay},
\texttt{Adware},
\texttt{Benign},
\texttt{Downloader},
\texttt{Trojan}
        },
        xtick={0,...,4},
        xtick style={draw=none},
	xticklabel style={anchor=east,rotate=30,yshift=-5pt,font=\tt\large},
        yticklabels={
\texttt{AdDisplay},
\texttt{Adware},
\texttt{Benign},
\texttt{Downloader},
\texttt{Trojan}
        },
        ytick={0,...,4},
        ytick style={draw=none},
        enlargelimits=false,
        yticklabel style={font=\tt\large},
        colorbar,
        colorbar style={
            ytick={0.0,0.2,0.4,0.6,0.8,1.0},
            yticklabels={0.0,0.2,0.4,0.6,0.8,1.0},
            yticklabel={\pgfmathprintnumber\tick},
            yticklabel style={
            		scale=1.33,
            		/pgf/number format/fixed,
			/pgf/number format/precision=1}
        },
        point meta min=0.0,
        point meta max=1.0,
        nodes near coords={\pgfmathprintnumber\pgfplotspointmeta},
        nodes near coords black white/.style={
            small value/.style={
                yshift=-7pt,
                text=black,
                /pgf/number format/fixed,
                /pgf/number format/precision=3,
                /pgf/number format/zerofill=true,
                scale=1.2,
            },
            large value/.style={
                yshift=-7pt,
                text=white,
                /pgf/number format/fixed,
                /pgf/number format/precision=3,
                /pgf/number format/zerofill=true,
                scale=1.2,
            },
            every node near coord/.style={
                check for zero/.code={
                    \pgfmathfloatifflags{\pgfplotspointmeta}{0}{
                        \pgfkeys{/tikz/coordinate}
                    }{
                        \begingroup
                        \pgfkeys{/pgf/fpu}
                        \pgfmathparse{\pgfplotspointmeta<#1}
                        \global\let\result=\pgfmathresult
                        \endgroup
                        %
                        %
                        \pgfmathfloatcreate{1}{1.0}{0}
                        \let\ONE=\pgfmathresult
                        \ifx\result\ONE
                            \pgfkeysalso{/pgfplots/small value}
                        \else
                            \pgfkeysalso{/pgfplots/large value}
                        \fi
                    }
                },
                check for zero,
            },
        },
        nodes near coords black white=0.5,
    ]
        \addplot[
            matrix plot,
            mesh/cols=5,
            point meta=explicit,draw=gray
        ] table [meta=C] {
            x y C
0 0 0.70
1 0 0.10
2 0 0.19
3 0 0.00
4 0 0.01
0 1 0.00
1 1 0.88
2 1 0.12
3 1 0.00
4 1 0.00
0 2 0.19
1 2 0.17
2 2 0.61
3 2 0.00
4 2 0.03
0 3 0.00
1 3 0.00
2 3 0.00
3 3 0.99
4 3 0.01
0 4 0.05
1 4 0.05
2 4 0.03
3 4 0.02
4 4 0.85
         };
    \end{axis}
\end{tikzpicture}
&
\begin{tikzpicture}[scale=0.4]
    \begin{axis}[
        width=10cm,
        height=10cm,
	colormap={bluewhite}{color=(white) rgb255=(100,149,237)},
        xticklabels={
\texttt{AdDisplay},
\texttt{Adware},
\texttt{Benign},
\texttt{Downloader},
\texttt{Trojan}
        },
        xtick={0,...,4},
        xtick style={draw=none},
	xticklabel style={anchor=east,rotate=30,yshift=-5pt,font=\tt\large},
        yticklabels={
\texttt{AdDisplay},
\texttt{Adware},
\texttt{Benign},
\texttt{Downloader},
\texttt{Trojan}
        },
        ytick={0,...,4},
        ytick style={draw=none},
        enlargelimits=false,
        yticklabel style={font=\tt\large},
        colorbar,
        colorbar style={
            ytick={0.0,0.2,0.4,0.6,0.8,1.0},
            yticklabels={0.0,0.2,0.4,0.6,0.8,1.0},
            yticklabel={\pgfmathprintnumber\tick},
            yticklabel style={
            		scale=1.33,
            		/pgf/number format/fixed,
			/pgf/number format/precision=1}
        },
        point meta min=0.0,
        point meta max=1.0,
        nodes near coords={\pgfmathprintnumber\pgfplotspointmeta},
        nodes near coords black white/.style={
            small value/.style={
                yshift=-7pt,
                text=black,
                /pgf/number format/fixed,
                /pgf/number format/precision=3,
                /pgf/number format/zerofill=true,
                scale=1.2,
            },
            large value/.style={
                yshift=-7pt,
                text=white,
                /pgf/number format/fixed,
                /pgf/number format/precision=3,
                /pgf/number format/zerofill=true,
                scale=1.2,
            },
            every node near coord/.style={
                check for zero/.code={
                    \pgfmathfloatifflags{\pgfplotspointmeta}{0}{
                        \pgfkeys{/tikz/coordinate}
                    }{
                        \begingroup
                        \pgfkeys{/pgf/fpu}
                        \pgfmathparse{\pgfplotspointmeta<#1}
                        \global\let\result=\pgfmathresult
                        \endgroup
                        %
                        %
                        \pgfmathfloatcreate{1}{1.0}{0}
                        \let\ONE=\pgfmathresult
                        \ifx\result\ONE
                            \pgfkeysalso{/pgfplots/small value}
                        \else
                            \pgfkeysalso{/pgfplots/large value}
                        \fi
                    }
                },
                check for zero,
            },
        },
        nodes near coords black white=0.5,
    ]
        \addplot[
            matrix plot,
            mesh/cols=5,
            point meta=explicit,draw=gray
        ] table [meta=C] {
            x y C
0 0 0.665
1 0 0.10
2 0 0.155
3 0 0.01
4 0 0.07
0 1 0.055
1 1 0.84
2 1 0.085
3 1 0.00
4 1 0.02
0 2 0.22
1 2 0.295
2 2 0.375
3 2 0.005
4 2 0.095
0 3 0.00
1 3 0.00
2 3 0.00
3 3 0.99
4 3 0.01
0 4 0.09
1 4 0.16
2 4 0.03
3 4 0.03
4 4 0.69
         };
    \end{axis}
\end{tikzpicture}
\\[-1ex]
\adjustbox{scale=0.85}{(a) MLP}
&
\adjustbox{scale=0.85}{(b) WL-Kernel}
\\
\\[-2ex]
\begin{tikzpicture}[scale=0.4]
    \begin{axis}[
        width=10cm,
        height=10cm,
	colormap={bluewhite}{color=(white) rgb255=(100,149,237)},
        xticklabels={
\texttt{AdDisplay},
\texttt{Adware},
\texttt{Benign},
\texttt{Downloader},
\texttt{Trojan}
        },
        xtick={0,...,4},
        xtick style={draw=none},
	xticklabel style={anchor=east,rotate=30,yshift=-5pt,font=\tt\large},
        yticklabels={
\texttt{AdDisplay},
\texttt{Adware},
\texttt{Benign},
\texttt{Downloader},
\texttt{Trojan}
        },
        ytick={0,...,4},
        ytick style={draw=none},
        enlargelimits=false,
        yticklabel style={font=\tt\large},
        colorbar,
        colorbar style={
            ytick={0.0,0.2,0.4,0.6,0.8,1.0},
            yticklabels={0.0,0.2,0.4,0.6,0.8,1.0},
            yticklabel={\pgfmathprintnumber\tick},
            yticklabel style={
            		scale=1.33,
            		/pgf/number format/fixed,
			/pgf/number format/precision=1}
        },
        point meta min=0.0,
        point meta max=1.0,
        nodes near coords={\pgfmathprintnumber\pgfplotspointmeta},
        nodes near coords black white/.style={
            small value/.style={
                yshift=-7pt,
                text=black,
                /pgf/number format/fixed,
                /pgf/number format/precision=3,
                /pgf/number format/zerofill=true,
                scale=1.2,
            },
            large value/.style={
                yshift=-7pt,
                text=white,
                /pgf/number format/fixed,
                /pgf/number format/precision=3,
                /pgf/number format/zerofill=true,
                scale=1.2,
            },
            every node near coord/.style={
                check for zero/.code={
                    \pgfmathfloatifflags{\pgfplotspointmeta}{0}{
                        \pgfkeys{/tikz/coordinate}
                    }{
                        \begingroup
                        \pgfkeys{/pgf/fpu}
                        \pgfmathparse{\pgfplotspointmeta<#1}
                        \global\let\result=\pgfmathresult
                        \endgroup
                        %
                        %
                        \pgfmathfloatcreate{1}{1.0}{0}
                        \let\ONE=\pgfmathresult
                        \ifx\result\ONE
                            \pgfkeysalso{/pgfplots/small value}
                        \else
                            \pgfkeysalso{/pgfplots/large value}
                        \fi
                    }
                },
                check for zero,
            },
        },
        nodes near coords black white=0.5,
    ]
        \addplot[
            matrix plot,
            mesh/cols=5,
            point meta=explicit,draw=gray
        ] table [meta=C] {
            x y C
0 0 0.86
1 0 0.02
2 0 0.11
3 0 0.00
4 0 0.01
0 1 0.015
1 1 0.885
2 1 0.08
3 1 0.00
4 1 0.02
0 2 0.145
1 2 0.13
2 2 0.71
3 2 0.00
4 2 0.015
0 3 0.00
1 3 0.00
2 3 0.00
3 3 0.995
4 3 0.005
0 4 0.04
1 4 0.085
2 4 0.065
3 4 0.015
4 4 0.795
         };
    \end{axis}
\end{tikzpicture}
&
\begin{tikzpicture}[scale=0.4]
    \begin{axis}[
        width=10cm,
        height=10cm,
	colormap={bluewhite}{color=(white) rgb255=(100,149,237)},
        xticklabels={
\texttt{AdDisplay},
\texttt{Adware},
\texttt{Benign},
\texttt{Downloader},
\texttt{Trojan}
        },
        xtick={0,...,4},
        xtick style={draw=none},
	xticklabel style={anchor=east,rotate=30,yshift=-5pt,font=\tt\large},
        yticklabels={
\texttt{AdDisplay},
\texttt{Adware},
\texttt{Benign},
\texttt{Downloader},
\texttt{Trojan}
        },
        ytick={0,...,4},
        ytick style={draw=none},
        enlargelimits=false,
        yticklabel style={font=\tt\large},
        colorbar,
        colorbar style={
            ytick={0.0,0.2,0.4,0.6,0.8,1.0},
            yticklabels={0.0,0.2,0.4,0.6,0.8,1.0},
            yticklabel={\pgfmathprintnumber\tick},
            yticklabel style={
            		scale=1.33,
            		/pgf/number format/fixed,
			/pgf/number format/precision=1}
        },
        point meta min=0.0,
        point meta max=1.0,
        nodes near coords={\pgfmathprintnumber\pgfplotspointmeta},
        nodes near coords black white/.style={
            small value/.style={
                yshift=-7pt,
                text=black,
                /pgf/number format/fixed,
                /pgf/number format/precision=3,
                /pgf/number format/zerofill=true,
                scale=1.2,
            },
            large value/.style={
                yshift=-7pt,
                text=white,
                /pgf/number format/fixed,
                /pgf/number format/precision=3,
                /pgf/number format/zerofill=true,
                scale=1.2,
            },
            every node near coord/.style={
                check for zero/.code={
                    \pgfmathfloatifflags{\pgfplotspointmeta}{0}{
                        \pgfkeys{/tikz/coordinate}
                    }{
                        \begingroup
                        \pgfkeys{/pgf/fpu}
                        \pgfmathparse{\pgfplotspointmeta<#1}
                        \global\let\result=\pgfmathresult
                        \endgroup
                        %
                        %
                        \pgfmathfloatcreate{1}{1.0}{0}
                        \let\ONE=\pgfmathresult
                        \ifx\result\ONE
                            \pgfkeysalso{/pgfplots/small value}
                        \else
                            \pgfkeysalso{/pgfplots/large value}
                        \fi
                    }
                },
                check for zero,
            },
        },
        nodes near coords black white=0.5,
    ]
        \addplot[
            matrix plot,
            mesh/cols=5,
            point meta=explicit,draw=gray
        ] table [meta=C] {
            x y C
0 0 0.80
1 0 0.06
2 0 0.13
3 0 0.00
4 0 0.01
0 1 0.04
1 1 0.765
2 1 0.15
3 1 0.00
4 1 0.045
0 2 0.175
1 2 0.145
2 2 0.615
3 2 0.00
4 2 0.065
0 3 0.00
1 3 0.00
2 3 0.005
3 3 0.99
4 3 0.005
0 4 0.035
1 4 0.13
2 4 0.08
3 4 0.025
4 4 0.73
         };
    \end{axis}
\end{tikzpicture}
\\[-1ex]
\adjustbox{scale=0.85}{(c) FEATHER}
&
\adjustbox{scale=0.85}{(d) Slaq-LSD}
\\
\\[-2ex]
\multicolumn{2}{c}{
\begin{tikzpicture}[scale=0.4]
    \begin{axis}[
        width=10cm,
        height=10cm,
	colormap={bluewhite}{color=(white) rgb255=(100,149,237)},
        xticklabels={
\texttt{AdDisplay},
\texttt{Adware},
\texttt{Benign},
\texttt{Downloader},
\texttt{Trojan}
        },
        xtick={0,...,4},
        xtick style={draw=none},
	xticklabel style={anchor=east,rotate=30,yshift=-5pt,font=\tt\large},
        yticklabels={
\texttt{AdDisplay},
\texttt{Adware},
\texttt{Benign},
\texttt{Downloader},
\texttt{Trojan}
        },
        ytick={0,...,4},
        ytick style={draw=none},
        enlargelimits=false,
        yticklabel style={font=\tt\large},
        colorbar,
        colorbar style={
            ytick={0.0,0.2,0.4,0.6,0.8,1.0},
            yticklabels={0.0,0.2,0.4,0.6,0.8,1.0},
            yticklabel={\pgfmathprintnumber\tick},
            yticklabel style={
            		scale=1.33,
            		/pgf/number format/fixed,
			/pgf/number format/precision=1}
        },
        point meta min=0.0,
        point meta max=1.0,
        nodes near coords={\pgfmathprintnumber\pgfplotspointmeta},
        nodes near coords black white/.style={
            small value/.style={
                yshift=-7pt,
                text=black,
                /pgf/number format/fixed,
                /pgf/number format/precision=3,
                /pgf/number format/zerofill=true,
                scale=1.2,
            },
            large value/.style={
                yshift=-7pt,
                text=white,
                /pgf/number format/fixed,
                /pgf/number format/precision=3,
                /pgf/number format/zerofill=true,
                scale=1.2,
            },
            every node near coord/.style={
                check for zero/.code={
                    \pgfmathfloatifflags{\pgfplotspointmeta}{0}{
                        \pgfkeys{/tikz/coordinate}
                    }{
                        \begingroup
                        \pgfkeys{/pgf/fpu}
                        \pgfmathparse{\pgfplotspointmeta<#1}
                        \global\let\result=\pgfmathresult
                        \endgroup
                        %
                        %
                        \pgfmathfloatcreate{1}{1.0}{0}
                        \let\ONE=\pgfmathresult
                        \ifx\result\ONE
                            \pgfkeysalso{/pgfplots/small value}
                        \else
                            \pgfkeysalso{/pgfplots/large value}
                        \fi
                    }
                },
                check for zero,
            },
        },
        nodes near coords black white=0.5,
    ]
        \addplot[
            matrix plot,
            mesh/cols=5,
            point meta=explicit,draw=gray
        ] table [meta=C] {
            x y C
0 0 0.56
1 0 0.28
2 0 0.135
3 0 0.00
4 0 0.025
0 1 0.33
1 1 0.55
2 1 0.115
3 1 0.00
4 1 0.005
0 2 0.225
1 2 0.47
2 2 0.21
3 2 0.005
4 2 0.09
0 3 0.00
1 3 0.00
2 3 0.005
3 3 0.92
4 3 0.075
0 4 0.11
1 4 0.215
2 4 0.095
3 4 0.055
4 4 0.525
         };
    \end{axis}
\end{tikzpicture}}
\\[-1ex]
\multicolumn{2}{c}{\adjustbox{scale=0.85}{(e) Slaq-VNGE}}
\end{tabular}
\caption{Confusion matrices for non-GNN models}\label{fig:conf_non}
\end{figure}

\begin{figure}[!htb]
\centering
\begin{tabular}{cc}
\begin{tikzpicture}[scale=0.4]
    \begin{axis}[
        width=10cm,
        height=10cm,
	colormap={bluewhite}{color=(white) rgb255=(100,149,237)},
        xticklabels={
\texttt{AdDisplay},
\texttt{Adware},
\texttt{Benign},
\texttt{Downloader},
\texttt{Trojan}
        },
        xtick={0,...,4},
        xtick style={draw=none},
	xticklabel style={anchor=east,rotate=30,yshift=-5pt,font=\tt\large},
        yticklabels={
\texttt{AdDisplay},
\texttt{Adware},
\texttt{Benign},
\texttt{Downloader},
\texttt{Trojan}
        },
        ytick={0,...,4},
        ytick style={draw=none},
        enlargelimits=false,
        yticklabel style={font=\tt\large},
        colorbar,
        colorbar style={
            ytick={0.0,0.2,0.4,0.6,0.8,1.0},
            yticklabels={0.0,0.2,0.4,0.6,0.8,1.0},
            yticklabel={\pgfmathprintnumber\tick},
            yticklabel style={
            		scale=1.33,
            		/pgf/number format/fixed,
			/pgf/number format/precision=1}
        },
        point meta min=0.0,
        point meta max=1.0,
        nodes near coords={\pgfmathprintnumber\pgfplotspointmeta},
        nodes near coords black white/.style={
            small value/.style={
                yshift=-7pt,
                text=black,
                /pgf/number format/fixed,
                /pgf/number format/precision=3,
                /pgf/number format/zerofill=true,
                scale=1.2,
            },
            large value/.style={
                yshift=-7pt,
                text=white,
                /pgf/number format/fixed,
                /pgf/number format/precision=3,
                /pgf/number format/zerofill=true,
                scale=1.2,
            },
            every node near coord/.style={
                check for zero/.code={
                    \pgfmathfloatifflags{\pgfplotspointmeta}{0}{
                        \pgfkeys{/tikz/coordinate}
                    }{
                        \begingroup
                        \pgfkeys{/pgf/fpu}
                        \pgfmathparse{\pgfplotspointmeta<#1}
                        \global\let\result=\pgfmathresult
                        \endgroup
                        %
                        %
                        \pgfmathfloatcreate{1}{1.0}{0}
                        \let\ONE=\pgfmathresult
                        \ifx\result\ONE
                            \pgfkeysalso{/pgfplots/small value}
                        \else
                            \pgfkeysalso{/pgfplots/large value}
                        \fi
                    }
                },
                check for zero,
            },
        },
        nodes near coords black white=0.5,
    ]
        \addplot[
            matrix plot,
            mesh/cols=5,
            point meta=explicit,draw=gray
        ] table [meta=C] {
            x y C
0 0 0.98
1 0 0.00
2 0 0.02
3 0 0.00
4 0 0.00
0 1 0.00
1 1 0.94
2 1 0.04
3 1 0.00
4 1 0.02
0 2 0.00
1 2 0.03
2 2 0.95
3 2 0.00
4 2 0.02
0 3 0.00
1 3 0.00
2 3 0.00
3 3 0.99
4 3 0.01
0 4 0.00
1 4 0.00
2 4 0.05
3 4 0.02
4 4 0.93
         };
    \end{axis}
\end{tikzpicture}
&
\begin{tikzpicture}[scale=0.4]
    \begin{axis}[
        width=10cm,
        height=10cm,
	colormap={bluewhite}{color=(white) rgb255=(100,149,237)},
        xticklabels={
\texttt{AdDisplay},
\texttt{Adware},
\texttt{Benign},
\texttt{Downloader},
\texttt{Trojan}
        },
        xtick={0,...,4},
        xtick style={draw=none},
	xticklabel style={anchor=east,rotate=30,yshift=-5pt,font=\tt\large},
        yticklabels={
\texttt{AdDisplay},
\texttt{Adware},
\texttt{Benign},
\texttt{Downloader},
\texttt{Trojan}
        },
        ytick={0,...,4},
        ytick style={draw=none},
        enlargelimits=false,
        yticklabel style={font=\tt\large},
        colorbar,
        colorbar style={
            ytick={0.0,0.2,0.4,0.6,0.8,1.0},
            yticklabels={0.0,0.2,0.4,0.6,0.8,1.0},
            yticklabel={\pgfmathprintnumber\tick},
            yticklabel style={
            		scale=1.33,
            		/pgf/number format/fixed,
			/pgf/number format/precision=1}
        },
        point meta min=0.0,
        point meta max=1.0,
        nodes near coords={\pgfmathprintnumber\pgfplotspointmeta},
        nodes near coords black white/.style={
            small value/.style={
                yshift=-7pt,
                text=black,
                /pgf/number format/fixed,
                /pgf/number format/precision=3,
                /pgf/number format/zerofill=true,
                scale=1.2,
            },
            large value/.style={
                yshift=-7pt,
                text=white,
                /pgf/number format/fixed,
                /pgf/number format/precision=3,
                /pgf/number format/zerofill=true,
                scale=1.2,
            },
            every node near coord/.style={
                check for zero/.code={
                    \pgfmathfloatifflags{\pgfplotspointmeta}{0}{
                        \pgfkeys{/tikz/coordinate}
                    }{
                        \begingroup
                        \pgfkeys{/pgf/fpu}
                        \pgfmathparse{\pgfplotspointmeta<#1}
                        \global\let\result=\pgfmathresult
                        \endgroup
                        %
                        %
                        \pgfmathfloatcreate{1}{1.0}{0}
                        \let\ONE=\pgfmathresult
                        \ifx\result\ONE
                            \pgfkeysalso{/pgfplots/small value}
                        \else
                            \pgfkeysalso{/pgfplots/large value}
                        \fi
                    }
                },
                check for zero,
            },
        },
        nodes near coords black white=0.5,
    ]
        \addplot[
            matrix plot,
            mesh/cols=5,
            point meta=explicit,draw=gray
        ] table [meta=C] {
            x y C
0 0 0.95
1 0 0.00
2 0 0.02
3 0 0.00
4 0 0.03
0 1 0.02
1 1 0.56
2 1 0.39
3 1 0.00
4 1 0.03
0 2 0.16
1 2 0.00
2 2 0.59
3 2 0.00
4 2 0.24
0 3 0.00
1 3 0.00
2 3 0.00
3 3 1.00
4 3 0.00
0 4 0.01
1 4 0.00
2 4 0.06
3 4 0.05
4 4 0.88
         };
    \end{axis}
\end{tikzpicture}
\\[-1ex]
\adjustbox{scale=0.85}{(a) GCN}
&
\adjustbox{scale=0.85}{(b) GraphSAGE}
\\
\\[-2ex]
\begin{tikzpicture}[scale=0.4]
    \begin{axis}[
        width=10cm,
        height=10cm,
	colormap={bluewhite}{color=(white) rgb255=(100,149,237)},
        xticklabels={
\texttt{AdDisplay},
\texttt{Adware},
\texttt{Benign},
\texttt{Downloader},
\texttt{Trojan}
        },
        xtick={0,...,4},
        xtick style={draw=none},
	xticklabel style={anchor=east,rotate=30,yshift=-5pt,font=\tt\large},
        yticklabels={
\texttt{AdDisplay},
\texttt{Adware},
\texttt{Benign},
\texttt{Downloader},
\texttt{Trojan}
        },
        ytick={0,...,4},
        ytick style={draw=none},
        enlargelimits=false,
        yticklabel style={font=\tt\large},
        colorbar,
        colorbar style={
            ytick={0.0,0.2,0.4,0.6,0.8,1.0},
            yticklabels={0.0,0.2,0.4,0.6,0.8,1.0},
            yticklabel={\pgfmathprintnumber\tick},
            yticklabel style={
            		scale=1.33,
            		/pgf/number format/fixed,
			/pgf/number format/precision=1}
        },
        point meta min=0.0,
        point meta max=1.0,
        nodes near coords={\pgfmathprintnumber\pgfplotspointmeta},
        nodes near coords black white/.style={
            small value/.style={
                yshift=-7pt,
                text=black,
                /pgf/number format/fixed,
                /pgf/number format/precision=3,
                /pgf/number format/zerofill=true,
                scale=1.2,
            },
            large value/.style={
                yshift=-7pt,
                text=white,
                /pgf/number format/fixed,
                /pgf/number format/precision=3,
                /pgf/number format/zerofill=true,
                scale=1.2,
            },
            every node near coord/.style={
                check for zero/.code={
                    \pgfmathfloatifflags{\pgfplotspointmeta}{0}{
                        \pgfkeys{/tikz/coordinate}
                    }{
                        \begingroup
                        \pgfkeys{/pgf/fpu}
                        \pgfmathparse{\pgfplotspointmeta<#1}
                        \global\let\result=\pgfmathresult
                        \endgroup
                        %
                        %
                        \pgfmathfloatcreate{1}{1.0}{0}
                        \let\ONE=\pgfmathresult
                        \ifx\result\ONE
                            \pgfkeysalso{/pgfplots/small value}
                        \else
                            \pgfkeysalso{/pgfplots/large value}
                        \fi
                    }
                },
                check for zero,
            },
        },
        nodes near coords black white=0.5,
    ]
        \addplot[
            matrix plot,
            mesh/cols=5,
            point meta=explicit,draw=gray
        ] table [meta=C] {
            x y C
0 0 0.95
1 0 0.00
2 0 0.05
3 0 0.00
4 0 0.00
0 1 0.00
1 1 0.90
2 1 0.09
3 1 0.00
4 1 0.01
0 2 0.00
1 2 0.01
2 2 0.99
3 2 0.00
4 2 0.00
0 3 0.00
1 3 0.00
2 3 0.00
3 3 1.00
4 3 0.00
0 4 0.00
1 4 0.01
2 4 0.09
3 4 0.04
4 4 0.86
         };
    \end{axis}
\end{tikzpicture}
&
\begin{tikzpicture}[scale=0.4]
    \begin{axis}[
        width=10cm,
        height=10cm,
	colormap={bluewhite}{color=(white) rgb255=(100,149,237)},
        xticklabels={
\texttt{AdDisplay},
\texttt{Adware},
\texttt{Benign},
\texttt{Downloader},
\texttt{Trojan}
        },
        xtick={0,...,4},
        xtick style={draw=none},
	xticklabel style={anchor=east,rotate=30,yshift=-5pt,font=\tt\large},
        yticklabels={
\texttt{AdDisplay},
\texttt{Adware},
\texttt{Benign},
\texttt{Downloader},
\texttt{Trojan}
        },
        ytick={0,...,4},
        ytick style={draw=none},
        enlargelimits=false,
        yticklabel style={font=\tt\large},
        colorbar,
        colorbar style={
            ytick={0.0,0.2,0.4,0.6,0.8,1.0},
            yticklabels={0.0,0.2,0.4,0.6,0.8,1.0},
            yticklabel={\pgfmathprintnumber\tick},
            yticklabel style={
            		scale=1.33,
            		/pgf/number format/fixed,
			/pgf/number format/precision=1}
        },
        point meta min=0.0,
        point meta max=1.0,
        nodes near coords={\pgfmathprintnumber\pgfplotspointmeta},
        nodes near coords black white/.style={
            small value/.style={
                yshift=-7pt,
                text=black,
                /pgf/number format/fixed,
                /pgf/number format/precision=3,
                /pgf/number format/zerofill=true,
                scale=1.2,
            },
            large value/.style={
                yshift=-7pt,
                text=white,
                /pgf/number format/fixed,
                /pgf/number format/precision=3,
                /pgf/number format/zerofill=true,
                scale=1.2,
            },
            every node near coord/.style={
                check for zero/.code={
                    \pgfmathfloatifflags{\pgfplotspointmeta}{0}{
                        \pgfkeys{/tikz/coordinate}
                    }{
                        \begingroup
                        \pgfkeys{/pgf/fpu}
                        \pgfmathparse{\pgfplotspointmeta<#1}
                        \global\let\result=\pgfmathresult
                        \endgroup
                        %
                        %
                        \pgfmathfloatcreate{1}{1.0}{0}
                        \let\ONE=\pgfmathresult
                        \ifx\result\ONE
                            \pgfkeysalso{/pgfplots/small value}
                        \else
                            \pgfkeysalso{/pgfplots/large value}
                        \fi
                    }
                },
                check for zero,
            },
        },
        nodes near coords black white=0.5,
    ]
        \addplot[
            matrix plot,
            mesh/cols=5,
            point meta=explicit,draw=gray
        ] table [meta=C] {
            x y C
0 0 0.91
1 0 0.00
2 0 0.09
3 0 0.00
4 0 0.00
0 1 0.00
1 1 0.88
2 1 0.11
3 1 0.00
4 1 0.01
0 2 0.10
1 2 0.00
2 2 0.89
3 2 0.00
4 2 0.01
0 3 0.00
1 3 0.00
2 3 0.00
3 3 0.99
4 3 0.01
0 4 0.00
1 4 0.00
2 4 0.10
3 4 0.04
4 4 0.86
         };
    \end{axis}
\end{tikzpicture}
\\[-1ex]
\adjustbox{scale=0.85}{(c) GIN}
&
\adjustbox{scale=0.85}{(d) SGC}
\\
\\[-2ex]
\begin{tikzpicture}[scale=0.4]
    \begin{axis}[
        width=10cm,
        height=10cm,
	colormap={bluewhite}{color=(white) rgb255=(100,149,237)},
        xticklabels={
\texttt{AdDisplay},
\texttt{Adware},
\texttt{Benign},
\texttt{Downloader},
\texttt{Trojan}
        },
        xtick={0,...,4},
        xtick style={draw=none},
	xticklabel style={anchor=east,rotate=30,yshift=-5pt,font=\tt\large},
        yticklabels={
\texttt{AdDisplay},
\texttt{Adware},
\texttt{Benign},
\texttt{Downloader},
\texttt{Trojan}
        },
        ytick={0,...,4},
        ytick style={draw=none},
        enlargelimits=false,
        yticklabel style={font=\tt\large},
        colorbar,
        colorbar style={
            ytick={0.0,0.2,0.4,0.6,0.8,1.0},
            yticklabels={0.0,0.2,0.4,0.6,0.8,1.0},
            yticklabel={\pgfmathprintnumber\tick},
            yticklabel style={
            		scale=1.33,
            		/pgf/number format/fixed,
			/pgf/number format/precision=1}
        },
        point meta min=0.0,
        point meta max=1.0,
        nodes near coords={\pgfmathprintnumber\pgfplotspointmeta},
        nodes near coords black white/.style={
            small value/.style={
                yshift=-7pt,
                text=black,
                /pgf/number format/fixed,
                /pgf/number format/precision=3,
                /pgf/number format/zerofill=true,
                scale=1.2,
            },
            large value/.style={
                yshift=-7pt,
                text=white,
                /pgf/number format/fixed,
                /pgf/number format/precision=3,
                /pgf/number format/zerofill=true,
                scale=1.2,
            },
            every node near coord/.style={
                check for zero/.code={
                    \pgfmathfloatifflags{\pgfplotspointmeta}{0}{
                        \pgfkeys{/tikz/coordinate}
                    }{
                        \begingroup
                        \pgfkeys{/pgf/fpu}
                        \pgfmathparse{\pgfplotspointmeta<#1}
                        \global\let\result=\pgfmathresult
                        \endgroup
                        %
                        %
                        \pgfmathfloatcreate{1}{1.0}{0}
                        \let\ONE=\pgfmathresult
                        \ifx\result\ONE
                            \pgfkeysalso{/pgfplots/small value}
                        \else
                            \pgfkeysalso{/pgfplots/large value}
                        \fi
                    }
                },
                check for zero,
            },
        },
        nodes near coords black white=0.5,
    ]
        \addplot[
            matrix plot,
            mesh/cols=5,
            point meta=explicit,draw=gray
        ] table [meta=C] {
            x y C
0 0 0.85
1 0 0.03
2 0 0.08
3 0 0.00
4 0 0.04
0 1 0.00
1 1 0.91
2 1 0.07
3 1 0.00
4 1 0.02
0 2 0.04
1 2 0.05
2 2 0.84
3 2 0.00
4 2 0.07
0 3 0.00
1 3 0.00
2 3 0.00
3 3 1.00
4 3 0.00
0 4 0.02
1 4 0.02
2 4 0.06
3 4 0.03
4 4 0.87
         };
    \end{axis}
\end{tikzpicture}
&
\begin{tikzpicture}[scale=0.4]
    \begin{axis}[
        width=10cm,
        height=10cm,
	colormap={bluewhite}{color=(white) rgb255=(100,149,237)},
        xticklabels={
\texttt{AdDisplay},
\texttt{Adware},
\texttt{Benign},
\texttt{Downloader},
\texttt{Trojan}
        },
        xtick={0,...,4},
        xtick style={draw=none},
	xticklabel style={anchor=east,rotate=30,yshift=-5pt,font=\tt\large},
        yticklabels={
\texttt{AdDisplay},
\texttt{Adware},
\texttt{Benign},
\texttt{Downloader},
\texttt{Trojan}
        },
        ytick={0,...,4},
        ytick style={draw=none},
        enlargelimits=false,
        yticklabel style={font=\tt\large},
        colorbar,
        colorbar style={
            ytick={0.0,0.2,0.4,0.6,0.8,1.0},
            yticklabels={0.0,0.2,0.4,0.6,0.8,1.0},
            yticklabel={\pgfmathprintnumber\tick},
            yticklabel style={
            		scale=1.33,
            		/pgf/number format/fixed,
			/pgf/number format/precision=1}
        },
        point meta min=0.0,
        point meta max=1.0,
        nodes near coords={\pgfmathprintnumber\pgfplotspointmeta},
        nodes near coords black white/.style={
            small value/.style={
                yshift=-7pt,
                text=black,
                /pgf/number format/fixed,
                /pgf/number format/precision=3,
                /pgf/number format/zerofill=true,
                scale=1.2,
            },
            large value/.style={
                yshift=-7pt,
                text=white,
                /pgf/number format/fixed,
                /pgf/number format/precision=3,
                /pgf/number format/zerofill=true,
                scale=1.2,
            },
            every node near coord/.style={
                check for zero/.code={
                    \pgfmathfloatifflags{\pgfplotspointmeta}{0}{
                        \pgfkeys{/tikz/coordinate}
                    }{
                        \begingroup
                        \pgfkeys{/pgf/fpu}
                        \pgfmathparse{\pgfplotspointmeta<#1}
                        \global\let\result=\pgfmathresult
                        \endgroup
                        %
                        %
                        \pgfmathfloatcreate{1}{1.0}{0}
                        \let\ONE=\pgfmathresult
                        \ifx\result\ONE
                            \pgfkeysalso{/pgfplots/small value}
                        \else
                            \pgfkeysalso{/pgfplots/large value}
                        \fi
                    }
                },
                check for zero,
            },
        },
        nodes near coords black white=0.5,
    ]
        \addplot[
            matrix plot,
            mesh/cols=5,
            point meta=explicit,draw=gray
        ] table [meta=C] {
            x y C
0 0 0.99
1 0 0.00
2 0 0.01
3 0 0.00
4 0 0.00
0 1 0.00
1 1 0.94
2 1 0.04
3 1 0.00
4 1 0.02
0 2 0.09
1 2 0.04
2 2 0.83
3 2 0.00
4 2 0.04
0 3 0.00
1 3 0.00
2 3 0.00
3 3 1.00
4 3 0.00
0 4 0.02
1 4 0.01
2 4 0.05
3 4 0.03
4 4 0.89
         };
    \end{axis}
\end{tikzpicture}
\\[-1ex]
\adjustbox{scale=0.85}{(e) JK-GCN}
&
\adjustbox{scale=0.85}{(f) JK-GraphSAGE}
\\
\\[-2ex]
\end{tabular}
{
\advance\tabcolsep by-5pt
\begin{tabular}{ccc}
\begin{tikzpicture}[scale=0.4]
    \begin{axis}[
        width=10cm,
        height=10cm,
	colormap={bluewhite}{color=(white) rgb255=(100,149,237)},
        xticklabels={
\texttt{AdDisplay},
\texttt{Adware},
\texttt{Benign},
\texttt{Downloader},
\texttt{Trojan}
        },
        xtick={0,...,4},
        xtick style={draw=none},
	xticklabel style={anchor=east,rotate=30,yshift=-5pt,font=\tt\large},
        yticklabels={
\texttt{AdDisplay},
\texttt{Adware},
\texttt{Benign},
\texttt{Downloader},
\texttt{Trojan}
        },
        ytick={0,...,4},
        ytick style={draw=none},
        enlargelimits=false,
        yticklabel style={font=\tt\large},
        colorbar,
        colorbar style={
            ytick={0.0,0.2,0.4,0.6,0.8,1.0},
            yticklabels={0.0,0.2,0.4,0.6,0.8,1.0},
            yticklabel={\pgfmathprintnumber\tick},
            yticklabel style={
            		scale=1.33,
            		/pgf/number format/fixed,
			/pgf/number format/precision=1}
        },
        point meta min=0.0,
        point meta max=1.0,
        nodes near coords={\pgfmathprintnumber\pgfplotspointmeta},
        nodes near coords black white/.style={
            small value/.style={
                yshift=-7pt,
                text=black,
                /pgf/number format/fixed,
                /pgf/number format/precision=3,
                /pgf/number format/zerofill=true,
                scale=1.2,
            },
            large value/.style={
                yshift=-7pt,
                text=white,
                /pgf/number format/fixed,
                /pgf/number format/precision=3,
                /pgf/number format/zerofill=true,
                scale=1.2,
            },
            every node near coord/.style={
                check for zero/.code={
                    \pgfmathfloatifflags{\pgfplotspointmeta}{0}{
                        \pgfkeys{/tikz/coordinate}
                    }{
                        \begingroup
                        \pgfkeys{/pgf/fpu}
                        \pgfmathparse{\pgfplotspointmeta<#1}
                        \global\let\result=\pgfmathresult
                        \endgroup
                        %
                        %
                        \pgfmathfloatcreate{1}{1.0}{0}
                        \let\ONE=\pgfmathresult
                        \ifx\result\ONE
                            \pgfkeysalso{/pgfplots/small value}
                        \else
                            \pgfkeysalso{/pgfplots/large value}
                        \fi
                    }
                },
                check for zero,
            },
        },
        nodes near coords black white=0.5,
    ]
        \addplot[
            matrix plot,
            mesh/cols=5,
            point meta=explicit,draw=gray
        ] table [meta=C] {
            x y C
0 0 0.99
1 0 0.00
2 0 0.01
3 0 0.00
4 0 0.00
0 1 0.00
1 1 0.95
2 1 0.04
3 1 0.00
4 1 0.01
0 2 0.02
1 2 0.00
2 2 0.96
3 2 0.00
4 2 0.02
0 3 0.00
1 3 0.00
2 3 0.00
3 3 1.00
4 3 0.00
0 4 0.00
1 4 0.00
2 4 0.02
3 4 0.04
4 4 0.94
         };
    \end{axis}
\end{tikzpicture}
&
\begin{tikzpicture}[scale=0.4]
    \begin{axis}[
        width=10cm,
        height=10cm,
	colormap={bluewhite}{color=(white) rgb255=(100,149,237)},
        xticklabels={
\texttt{AdDisplay},
\texttt{Adware},
\texttt{Benign},
\texttt{Downloader},
\texttt{Trojan}
        },
        xtick={0,...,4},
        xtick style={draw=none},
	xticklabel style={anchor=east,rotate=30,yshift=-5pt,font=\tt\large},
        yticklabels={
\texttt{AdDisplay},
\texttt{Adware},
\texttt{Benign},
\texttt{Downloader},
\texttt{Trojan}
        },
        ytick={0,...,4},
        ytick style={draw=none},
        enlargelimits=false,
        yticklabel style={font=\tt\large},
        colorbar,
        colorbar style={
            ytick={0.0,0.2,0.4,0.6,0.8,1.0},
            yticklabels={0.0,0.2,0.4,0.6,0.8,1.0},
            yticklabel={\pgfmathprintnumber\tick},
            yticklabel style={
            		scale=1.33,
            		/pgf/number format/fixed,
			/pgf/number format/precision=1}
        },
        point meta min=0.0,
        point meta max=1.0,
        nodes near coords={\pgfmathprintnumber\pgfplotspointmeta},
        nodes near coords black white/.style={
            small value/.style={
                yshift=-7pt,
                text=black,
                /pgf/number format/fixed,
                /pgf/number format/precision=3,
                /pgf/number format/zerofill=true,
                scale=1.2,
            },
            large value/.style={
                yshift=-7pt,
                text=white,
                /pgf/number format/fixed,
                /pgf/number format/precision=3,
                /pgf/number format/zerofill=true,
                scale=1.2,
            },
            every node near coord/.style={
                check for zero/.code={
                    \pgfmathfloatifflags{\pgfplotspointmeta}{0}{
                        \pgfkeys{/tikz/coordinate}
                    }{
                        \begingroup
                        \pgfkeys{/pgf/fpu}
                        \pgfmathparse{\pgfplotspointmeta<#1}
                        \global\let\result=\pgfmathresult
                        \endgroup
                        %
                        %
                        \pgfmathfloatcreate{1}{1.0}{0}
                        \let\ONE=\pgfmathresult
                        \ifx\result\ONE
                            \pgfkeysalso{/pgfplots/small value}
                        \else
                            \pgfkeysalso{/pgfplots/large value}
                        \fi
                    }
                },
                check for zero,
            },
        },
        nodes near coords black white=0.5,
    ]
        \addplot[
            matrix plot,
            mesh/cols=5,
            point meta=explicit,draw=gray
        ] table [meta=C] {
            x y C
0 0 0.98
1 0 0.00
2 0 0.02
3 0 0.00
4 0 0.00
0 1 0.00
1 1 0.94
2 1 0.04
3 1 0.00
4 1 0.02
0 2 0.00
1 2 0.04
2 2 0.94
3 2 0.00
4 2 0.02
0 3 0.00
1 3 0.00
2 3 0.00
3 3 0.99
4 3 0.01
0 4 0.00
1 4 0.01
2 4 0.03
3 4 0.02
4 4 0.94
         };
    \end{axis}
\end{tikzpicture}
&
\begin{tikzpicture}[scale=0.4]
    \begin{axis}[
        width=10cm,
        height=10cm,
	colormap={bluewhite}{color=(white) rgb255=(100,149,237)},
        xticklabels={
\texttt{AdDisplay},
\texttt{Adware},
\texttt{Benign},
\texttt{Downloader},
\texttt{Trojan}
        },
        xtick={0,...,4},
        xtick style={draw=none},
	xticklabel style={anchor=east,rotate=30,yshift=-5pt,font=\tt\large},
        yticklabels={
\texttt{AdDisplay},
\texttt{Adware},
\texttt{Benign},
\texttt{Downloader},
\texttt{Trojan}
        },
        ytick={0,...,4},
        ytick style={draw=none},
        enlargelimits=false,
        yticklabel style={font=\tt\large},
        colorbar,
        colorbar style={
            ytick={0.0,0.2,0.4,0.6,0.8,1.0},
            yticklabels={0.0,0.2,0.4,0.6,0.8,1.0},
            yticklabel={\pgfmathprintnumber\tick},
            yticklabel style={
            		scale=1.33,
            		/pgf/number format/fixed,
			/pgf/number format/precision=1}
        },
        point meta min=0.0,
        point meta max=1.0,
        nodes near coords={\pgfmathprintnumber\pgfplotspointmeta},
        nodes near coords black white/.style={
            small value/.style={
                yshift=-7pt,
                text=black,
                /pgf/number format/fixed,
                /pgf/number format/precision=3,
                /pgf/number format/zerofill=true,
                scale=1.2,
            },
            large value/.style={
                yshift=-7pt,
                text=white,
                /pgf/number format/fixed,
                /pgf/number format/precision=3,
                /pgf/number format/zerofill=true,
                scale=1.2,
            },
            every node near coord/.style={
                check for zero/.code={
                    \pgfmathfloatifflags{\pgfplotspointmeta}{0}{
                        \pgfkeys{/tikz/coordinate}
                    }{
                        \begingroup
                        \pgfkeys{/pgf/fpu}
                        \pgfmathparse{\pgfplotspointmeta<#1}
                        \global\let\result=\pgfmathresult
                        \endgroup
                        %
                        %
                        \pgfmathfloatcreate{1}{1.0}{0}
                        \let\ONE=\pgfmathresult
                        \ifx\result\ONE
                            \pgfkeysalso{/pgfplots/small value}
                        \else
                            \pgfkeysalso{/pgfplots/large value}
                        \fi
                    }
                },
                check for zero,
            },
        },
        nodes near coords black white=0.5,
    ]
        \addplot[
            matrix plot,
            mesh/cols=5,
            point meta=explicit,draw=gray
        ] table [meta=C] {
            x y C
0 0 0.95
1 0 0.00
2 0 0.04
3 0 0.00
4 0 0.01
0 1 0.02
1 1 0.90
2 1 0.07
3 1 0.00
4 1 0.01
0 2 0.11
1 2 0.03
2 2 0.83
3 2 0.00
4 2 0.03
0 3 0.00
1 3 0.00
2 3 0.00
3 3 0.99
4 3 0.01
0 4 0.01
1 4 0.00
2 4 0.03
3 4 0.02
4 4 0.94
         };
    \end{axis}
\end{tikzpicture}
\\[-1ex]
\adjustbox{scale=0.85}{(e) JK-GIN}
&
\adjustbox{scale=0.85}{(f) UnetGraph}
&
\adjustbox{scale=0.85}{(g) DGCNN}
\end{tabular}
}
\caption{Confusion matrices for GNN models}\label{fig:conf_GNN}
\end{figure}

\end{document}